%% file: main.tex
\documentclass[acmlarge,nonacm,dvipsnames]{acmart}
\AtBeginDocument{%
  }
\usepackage{xcolor}
\usepackage{amsfonts}
\usepackage{amsmath}
\usepackage{cuted,lipsum}
\usepackage[most]{tcolorbox}
\usepackage{subcaption}
\usepackage{booktabs}
\usepackage{pdflscape}
\usepackage{tabularx}
\usepackage{graphicx}
\usepackage{tablefootnote}
\usepackage{threeparttable}
\usepackage{hyperref}
\usepackage{multirow}
\usepackage{array}
\usepackage{placeins}
\usepackage{balance}
\usepackage{makecell}
\usepackage{soul}
\usepackage{adjustbox}

\copyrightyear{2018}
\acmYear{2018}
\acmDOI{XXXXXXX.XXXXXXX}

\acmJournal{POMACS}
\acmVolume{37}
\acmNumber{4}
\acmArticle{111}
\acmMonth{8}

\begin{document}

\title{Heart2Mind: Human-Centered Contestable Psychiatric Disorder Diagnosis System using Wearable ECG Monitors}

\author{Hung Nguyen}
\authornote{Corresponding author.}
\email{hung.ntt@unb.ca}
\orcid{0000-0002-6750-9536}
\affiliation{%
  \institution{University of New Brunswick}
  \city{Fredericton}
  \state{New Brunswick}
  \country{Canada}
}
\affiliation{%
  \institution{National Research Council Canada}
  \city{Fredericton}
  \state{New Brunswick}
  \country{Canada}
}

\author{Alireza Rahimi}
\orcid{0009-0006-6512-7303}
\affiliation{%
  \institution{University of New Brunswick}
  \city{Fredericton}
  \state{New Brunswick}
  \country{Canada}
}
\email{alireza.rahimi@unb.ca}

\author{Veronica Whitford}
\orcid{0000−0001−5681−9929}
\affiliation{%
  \institution{University of New Brunswick}
  \city{Fredericton}
  \state{New Brunswick}
  \country{Canada}
}
\email{veronica.whitford@unb.ca}

\author{Hélène Fournier}
\orcid{0000-0002-8470-3226}
\affiliation{%
  \institution{National Research Council Canada}
  \city{Moncton}
  \state{New Brunswick}
  \country{Canada}
}
\email{helene.fournier@nrc-cnrc.gc.ca}

\author{Irina Kondratova}
\orcid{0000-0002-5406-2309}
\affiliation{%
  \institution{National Research Council Canada}
  \city{Fredericton}
  \state{New Brunswick}
  \country{Canada}
}
\email{irina.kondratova@nrc-cnrc.gc.ca}

\author{René Richard}
\orcid{0000-0002-1342-6225}
\affiliation{%
  \institution{National Research Council Canada}
  \city{Fredericton}
  \state{New Brunswick}
  \country{Canada}
}\affiliation{%
  \institution{University of New Brunswick}
  \city{Fredericton}
  \state{New Brunswick}
  \country{Canada}
}
\email{rene.richard@nrc-cnrc.gc.ca}

\author{Hung Cao}
\orcid{0000-0002-0788-4377}
\affiliation{%
  \institution{University of New Brunswick}
  \city{Fredericton}
  \state{New Brunswick}
  \country{Canada}
}
\email{hcao3@unb.ca}

\renewcommand{\shortauthors}{Nguyen et al.}

\begin{abstract}
\input{sec/0_abstract}
\end{abstract}





\maketitle

\input{sec/1_intro}
\input{sec/2_rw}

\input{sec/3_method}

\input{sec/4_imp}

\input{sec/5_model_exp}
\input{sec/7_dcs}
\input{sec/8_cls}
\input{sec/9_appx}

\bibliographystyle{ACM-Reference-Format}
\bibliography{ref}
\end{document}

%% file: sec/0_abstract.tex
Psychiatric disorders affect millions globally, yet their diagnosis faces significant challenges in clinical practice due to subjective assessments and accessibility concerns, leading to potential delays in treatment. To help address this issue, we present Heart2Mind, a human-centered contestable psychiatric disorder diagnosis system using wearable electrocardiogram (ECG) monitors. Our approach leverages cardiac biomarkers, particularly heart rate variability (HRV) and R-R intervals (RRI) time series, as objective indicators of autonomic dysfunction in psychiatric conditions. The system comprises three key components: (1) a Cardiac Monitoring Interface (CMI) for real-time data acquisition from Polar H9/H10 devices; (2) a Multi-Scale Temporal-Frequency Transformer (MSTFT) that processes RRI time series through integrated time-frequency domain analysis; (3) a Contestable Diagnosis Interface (CDI) combining Self-Adversarial Explanations (SAEs) with contestable Large Language Models (LLMs). Our MSTFT achieves 91.7\% accuracy on the HRV-ACC dataset using leave-one-out cross-validation, outperforming state-of-the-art methods. SAEs successfully detect inconsistencies in model predictions by comparing attention-based and gradient-based explanations, while LLMs enable clinicians to validate correct predictions and contest erroneous ones. This work demonstrates the feasibility of combining wearable technology with Explainable Artificial Intelligence (XAI) and contestable LLMs to create a transparent, contestable system for psychiatric diagnosis that maintains clinical oversight while leveraging advanced AI capabilities.
Our implementation is publicly available at: \url{https://github.com/Analytics-Everywhere-Lab/heart2mind}.

%% file: sec/1_intro.tex
\section{Introduction}
Psychiatric disorders refer to a wide variety of mental conditions that significantly affect an individual's mood, behavior, thinking, and overall functioning. 
Some common types of psychiatric disorders include schizophrenia, bipolar disorder, depression, etc., which affect 1-3\% of the global population \cite{pedersen2014comprehensive,merikangas2007lifetime,mcgrath2008schizophrenia}.
More specifically, schizophrenia is a psychotic disorder characterized by a combination of positive symptoms (e.g., delusions, hallucinations, disorganized thinking) and negative symptoms (e.g., alogia, blunted affect). 
In contrast, bipolar disorder is a mood disorder characterized by extreme mood swings that include manic episodes (e.g., elevated mood, increased energy levels) and depressive episodes (e.g., low mood, loss of interest in activities). 
Bipolar disorder can, however, be accompanied by psychotic features (e.g., delusions, hallucinations) in some individuals during manic, depressive, or mixed episodes. 
These deficits can negatively impact functional independence and quality of life in those experiencing these disorders.

The concerning rise in psychiatric disorders and suicide rates among psychiatric patients requires prompt attention. 
Indeed, research has found that people with psychiatric disorders are at a significantly higher risk of suicide than the general population. Studies have shown that up to 90\% of individuals who commit suicide have a diagnosable psychiatric disorder at the time of their death \cite{braadvik2018suicide}.
The life expectancy of individuals with schizophrenia is reduced compared to that of the general population. Over the past decade, the mortality rate among these patients has increased by 2.6\% relative to the healthy population \cite{tasci2024novel}.
Although clinically distinct disorders, schizophrenia and bipolar disorder may involve a common neurodevelopmental basis, as evidenced by genetic overlap \cite{bramon2001common,lichtenstein2009common} and similar cognitive deficits (e.g., executive functioning, verbal fluency) \cite{simonsen2009effect}.
It is essential that people with mental health conditions receive appropriate treatment and support to help control their symptoms and lower suicide risk.
While diagnostic criteria for schizophrenia and bipolar disorder are well-established in the Diagnostic and Statistical Manual of Mental Disorders (DSM-5) \cite{who2007international} and International Classification of Diseases (ICD-11) \cite{american2013diagnostic}, the current diagnostic process relies heavily on self-report and clinical interviews, both of which may involve subjective elements. 
Although efforts are being made to minimize bias and enhance objectivity through observation and standardized assessment procedures (e.g., Positive and Negative Syndrome Scale; PANSS \cite{andreasen1995correlational}), the diagnosis process can be time-consuming and subject to potential errors based on the clinician's expertise and patient rapport. Moreover, while recent research has explored using blood biomarkers to diagnose psychiatric disorders, this approach requires costly laboratory testing, which may not be accessible in some areas of the world \cite{hill2024precision}.

Therefore, there is an urgent need for the automated, accurate, and effective detection of psychiatric disorders. 
Recent advancements in wearable technology offer a promising alternative. Heart rate variability (HRV) and R-R intervals (RRI), easily measured by consumer-grade devices, have emerged as promising biomarkers for psychiatric disorders, such as schizophrenia and bipolar disorder, as both involve cardiac autonomic dysregulation \cite{benjamin2021heart,hinde2021wearable,speer2020measuring,nguyen2025human}. 
HRV reflects the autonomic nervous system's function and has shown correlations with psychiatric disorder severity \cite{benarroch1993central,berntson1997heart,benjamin2021heart,zhang2022imbalance}.
The integration of artificial intelligence (AI) and deep learning (DL) with HRV analysis could enhance clinical diagnostics. 
However, previous attempts have faced limitations, such as focusing on heart rate and accelerometry instead of HRV or using less precise photoplethysmography (PPG) sensors \cite{hinde2021wearable,stautland2022p178}.
Moreover, the ``black-box'' nature of advanced AI techniques poses challenges for clinical acceptance.

The growing integration of AI in healthcare has prompted regulatory frameworks emphasizing not only transparency but also contestability, (i.e., the ability for users to question, intervene in, and correct AI decisions). Recent regulations such as the General Data Protection Regulation (GDPR) and the EU AI Act underscore this shift from mere explainability to active contestation. 
This regulatory trend is echoed globally, with frameworks like Canada's Directive on Automated Decision-Making \cite{board_board_2019}, Health Canada \cite{canada2025}, UK's Medicines and Healthcare products Regulatory Agency (MHRA) \cite{and_2024} guidelines, U.S Food and Drug Administration (FDA) regulations \cite{center_2025}, and the Montréal Declaration on Responsible AI \cite{de_2023}, all incorporating contestability principles through requirements for recourse mechanisms, human oversight, and appeal processes in AI-driven healthcare decisions.
Ethically, issues of responsibility and agency are paramount (i.e., who is accountable when an AI’s suggestion contributes to a decision, and how we ensure the clinician remains the ultimate decision-maker). Providing clinicians with the means to contest and override AI decisions is a safeguard for patient safety and aligns with the principle that automation should support rather than replace human judgment.

In response to these challenges, our research makes the following key contributions:
\begin{itemize}
    \item \textbf{We propose Heart2Mind, a human-centered psychiatric disorder detection system} that integrates wearable ECG monitoring with advanced AI diagnostics and human-AI collaborative decision-making mechanisms, enabling continuous, objective assessment of psychiatric conditions through cardiac biomarkers.
    \item \textbf{We introduce a novel Multi-Scale Temporal-Frequency Transformer (MSTFT)} in classifying psychiatric disorders from RRI time series by effectively integrating multi-scale temporal features and frequency-domain wavelet features through a transformer-based architecture with cross-attention fusion.
    \item \textbf{We advance the integration of contestable AI within XAI through Self-Adversarial Explanations (SAEs)} that identify discrepancies between attention-based and gradient-based explanations, successfully detecting potential model errors and unfaithfulness.
    \item \textbf{We develop a contestable Large Language Models (LLMs) system} that empowers clinicians to actively challenge and validate AI decisions through natural language interaction, thereby maintaining human oversight in critical healthcare decisions.
\end{itemize}

The remainder of this paper is organized as follows: Section~\ref{sec:rw} reviews related work in ECG-based psychiatric disorder detection, XAI applications in healthcare, and the evolution toward human-centered contestable AI systems. Section~\ref{sec:method} presents our methodology, detailing the Heart2Mind framework's architecture and the design principles of our contestable LLMs system. Section~\ref{sec:impl} describes the implementation of each component, including the Cardiac Monitoring Interface (CMI), Multi-Scale Temporal-Frequency Transformer (MSTFT), and Contestable Diagnosis Interface (CDI) with the contestable LLMs system. Section~\ref{sec:exp} presents comprehensive experiments evaluating both the diagnostic performance of our MSTFT model and the effectiveness of our contestable AI approach in validating correct predictions and identifying errors. Section~\ref{sec:disc} discusses the broader implications of our work, including the potential of wearable ECG devices in psychiatric care and the challenges in developing human-centered contestable AI systems for healthcare. Finally, Section~\ref{sec:conc} concludes the paper with a summary of contributions and directions for future research.
As a final introductory remark, it should be noted that detailed acronyms and definitions used in the paper are listed in Table~\ref{tab:acronyms}.

\begin{table}[ht!]
    \caption{Acronyms and Definitions}
    \centering
    \begin{adjustbox}{max width=\textwidth}
    \begin{tabular}{llll}
    \toprule
    \textbf{Term} & \textbf{Definition} & \textbf{Term} & \textbf{Definition} \\ 
    \midrule
    % Acronyms. Found in LaTeX source using regex search. Look for : \([A-Za-z0-9]+\)
    AI & Artificial Intelligence & IChi2 & Interactive Chi-square \\
    ANN & Artificial Neural Network & IMV & Iterative Majority Voting \\
    ANS & Autonomic Nervous System & LF & Low-Frequency \\
    BP & Blood Pressure & LIME & Local Interpretable Model-agnostic Explanations \\
    BD & Bipolar Disorder & LLM & Large Language Models \\
    BF & Blood Flow & LOOCV & Leave-One-Out Cross-Validation \\
    BLE & Bluetooth Low Energy & MDWT & Multilevel Discrete Wavelet Transform \\
    BPM & Beats Per Minute & MRI & Magnetic Resonance Imaging \\
    CAM & Class Activation Mapping & MRR & Mean R-R intervals \\
    CDI & Contestable Diagnosis Interface & MSTFT & Multi-scale Temporal-Frequency Transformer \\
    CMI & Cardiac Monitoring Interface & non-EUT & Non-euthymic \\
    CNN & Convolutional Neural Network & PANSS & Positive and Negative Syndrome Scale \\
    DL & Deep Learning & PD & Psychiatric Disorders \\
    DP & Depression & PPG & Photoplethysmography \\
    DTW & Dynamic Time Warping & RPA & Recurrence Plot Analysis \\
    EEG & Electroencephalography & RRI & R-R Interval \\
    EMD & Empirical Mode Decomposition & RMSSD & Root Mean Square of Successive R-R Interval Differences \\
    EUT & Euthymic & SAE & Self-Adversarial Explanation \\
    fMRI & functional Magnetic Resonance Imaging & SHAP & SHapley Additive exPlanations \\
    FKNN & Fine K-Nearest Neighbor & SZ & Schizophrenia \\
    FN & False Negative & TN & True Negative \\
    FP & False Positive & TP & True Positive \\
    GDPR & General Data Protection Regulation & TQWT & Tunable-Q Wavelet Transform \\
    HC & Healthy Control & VLF & Very Low Frequency \\
    HF & High-Frequency & VMD & Variational Mode Decomposition \\
    HR & Heart Rate & WSN & Wavelet Scattering Network \\
    HRV & Heart Rate Variability & XAI & Explainable AI \\
    NN & Neural Network & Contestable LLMs & Definition : Contestable AI + LLMs \\
    \bottomrule
    \end{tabular}
    \end{adjustbox}
    \label{tab:acronyms}
\end{table}

%% file: sec/2_rw.tex
\section{Related Work}\label{sec:rw}
This section provides an overview of recent AI techniques and systems in psychiatric disorder prediction, especially with wearable devices. We introduce a new direction of XAI going towards contestable AI. We also demonstrate previous attempts to define and apply human-centered contestable AI systems in the healthcare context.

\subsection{ECG and Psychiatric Disorder Detection}
Traditionally, biomarker-based detection of psychiatric disorders has centered on neurophysiological signals, most notably electroencephalography (EEG) \cite{rawat2025psyneuronet,montazeri2023application}, or biological data, including magnetic resonance imaging (MRI) \cite{jimenez2024machine,howes2023neuroimaging}, functional MRI (fMRI) \cite{wu2024advances,wang2024schizophrenia}, and genetic analysis \cite{owen2023genomic}. In contrast, the diagnostic value of peripheral cardiovascular markers, such as ECG and HRV, has only recently garnered attention in the context of psychiatric disorder detection. 
% HRV is the variation of heartbeat with respect to time. 
% It is affected by the ANS and might provide details about mental disorders. Research has revealed that meditation and relaxation can enhance HRV, reflecting a favorable effect on the heart and the brain \cite{rajendra2006heart,koh2022automated,khare2023ecgpsychnet}.
% TODO
One of the most crucial organs in the human body is the heart, and it is tightly connected to the brain through a variety of physiological and neurological processes.
For instance, psychiatric disorders often involve physiological dysregulation that can be captured via cardiovascular biomarkers \cite{koh2022automated,wang2025heart}.
Several ways of how the heart and brain communicate with each other have been explored as follows \cite{khare2023ecgpsychnet}:
\begin{itemize}
    \item \textbf{Autonomic nervous system (ANS):} Heart functions such as heart rate (HR), blood pressure (BP), and blood flow (BF) are controlled by the ANS. BP and HR go up during the sympathetic (``fight-or-flight'') response, which is controlled by the sympathetic nervous system, whereas the same parameters go down during the parasympathetic (``rest-and-digest'') response, which is controlled by the parasympathetic nervous system. The brain controls the ANS and can influence the heart's function through this pathway \cite{rajendra2006heart,silvani2016brain}. For example, in healthy individuals, higher HRV denotes greater flexibility and adaptability of the ANS. Conversely, psychological stress and emotional distress can directly suppress HRV by engaging sympathetic pathways and withdrawing parasympathetic influence \cite{wang2025heart}.
    \item \textbf{Stress and emotions:} Stress and emotions also affect the rhythm of heartbeats. As an example, when an individual feels anxiety or fear, the brain orders the heart to speed up and increases BP \cite{sgoifo2009inevitable}. Reductions in high-frequency (HF) power indicate decreased parasympathetic tone, and alterations in low-frequency (LF) power or the LF/HF ratio suggest sympathetic overactivity \cite{kim2018stress}. Similarly, sustained stress can alter the heart’s physiological properties over time, increasing the risk of heart disease \cite{rajendra2006heart,montano2009heart}. 
    \item \textbf{Heart–brain feedback loop:} In a feedback loop known as the ``heart–brain axis'', the heart also interacts with the brain. Serotonin and oxytocin are two neurotransmitters, and the hormones produced have an impact on mood and brain activity \cite{montano2009heart,koh2022automated}. This feedback loop may also impact the ANS, resulting in modifications to BP and HR.
\end{itemize}
The relationship between the brain and the heart is intricate and crucial to human physiology. Knowing how these two organs interact with one another can help to detect psychiatric disorders, anxiety, and other neurological conditions accurately and enhance mental health. Indeed, empirical research strongly support the role of ECG-derived measures in mental health assessment \cite{garbarino2014empatica,cella2018using,khare2023ecgpsychnet,tasci2024novel,koh2022automated,corponi2024does,ksikazek10hrv,wang2025heart,telangore2025novel,baydili2025artificial,nguyen2025human,ainunhusna2020bipolar,inoue2022development}.

In recent years, several computational approaches have been explored to detect psychiatric conditions from ECG signals. 
\citet{tasci2024novel} published a 12-lead ECG dataset (i.e., Psychiatry ECG dataset\hyperlink{fn:2}{\textsuperscript{[b]}}) and proposed a ternary pattern-based automatic classification model for distinguishing bipolar disorder, depression, schizophrenia, and healthy controls. Their approach involved multi-level discrete wavelet transform (MDWT) and ternary pattern-based feature extraction, followed by interactive Chi-square (IChi2) feature selection, classification using an artificial neural network (ANN), and Iterative Majority Voting (IMV) to enhance accuracy. Their proposed model achieved a classification accuracy of up to 96.25\% when using IMV. 
Using the same dataset, \citet{khare2023ecgpsychnet} proposed ECGPsychNet, an 
ECG-based multi-class optimized hybrid ensemble detection model. Their model employed empirical mode decomposition (EMD), variational mode decomposition (VMD), and tunable-Q wavelet transform (TQWT) to extract multi-level ECG features, which were then classified using an ensemble of optimizable classifiers, achieving the overall classification accuracy of 98.15\% with class-specific detection rates of about 98.96\% for schizophrenia and 95.12\% for bipolar disorder.
Building on and extending these two previous studies, \citet{telangore2025novel} proposed a wavelet scattering network (WSN) for feature extraction combined with a Fine K-Nearest Neighbor (FKNN) classifier. Their approach achieved an overall classification accuracy of 99.8\% using ten-fold cross-validation.

%=== Wearable ECG/HRV dataset
Beyond controlled hospital recordings, wearable devices capable of recording ECG/HRV data have opened new avenues for psychiatric disorder detection, including enabling continuous monitoring beyond traditional clinical environments \cite{baydili2025artificial}.
\citet{valenza2016predicting} analyzed HRV features, including time-domain (i.e., mean/standard deviation RRI, RMSSD (Root Mean Square of Successive RRI Differences), pNN50 (Percentage of successive RRI that differ by more than 50 ms)), frequency-domain (i.e., VLF/LF/HF peaks), and nonlinear metrics (i.e., Poincaré SD1/SD2, Recurrence Plot Analysis (RPA)) collected via the PSYCHE wearable t-shirt with integrated fabric electrodes and applied a SVM classifier to forecast mood transitions between euthymic (EUT) and non-euthymic (non-EUT) states with an average accuracy of 69\%, reaching up to 83.3\% in individual cases.
\citet{tiryaki2021real} proposed a real-time convolutional neural network (CNN)-based method for detecting ST-segment depression episodes using single-lead ECG signals from the VivaLNK Continuous ECG Recorder. They extracted PQ junction, QRS complex, and ST segment features from ECG signals to train a CNN classifier, achieving an accuracy of over 95\%, with sensitivity of 98\% and specificity of 91\% using the PhysioNet European ST-T Database\hyperlink{fn:1}{\textsuperscript{[a]}}.
\citet{cella2018using} used a multi-sensor wearable Emphatica E4 smartwatch \cite{garbarino2014empatica} (capturing HRV, electrodermal activity, and motion) worn for six days to study people with schizophrenia. They found significant HRV differences compared to health controls, where lower vagal tone correlated with greater symptom severity.
Also collecting data via the Empatica E4 smartwatch, \citet{corponi2024does} investigated HRV changes during acute episodes of bipolar disorder using a Bayesian hierarchical model applied to lnRMSSD (natural logarithm of the RMSSD). Their longitudinal analysis of HRV changes in people experiencing mania or depressive episodes revealed that lnRMSSD increased as symptoms improved, with a 95.18\% probability of positive direction. However, they found no significant difference in HRV trajectories between manic and depressive episodes.

%== no information of device and dataset
In other studies where the information on the used device and dataset was not specified, \citet{ainunhusna2020bipolar} classified people with bipolar disorder and health controls based on HRV features extracted from ECG recordings sampled at 250 Hz and analyzed with an SVM classifier. They employed time-domain HRV features (i.e., MHR, SDNN, and RMSSD), achieving the highest classification accuracy of 93.8\% using quadratic SVM.
\citet{zang2022end} proposed an end-to-end DL approach for automatic depression recognition using two-lead ECG signals and a one-dimensional convolutional neural network (1D-CNN). Their method eliminated manual feature extraction and selection, directly learning relevant features from raw ECG signals, achieving a classification accuracy of 93.96\%.
% sensitivity of 89.43\%, and specificity of 98.49\%.

While multi-day observation provides comprehensive data, it may be burdensome and often relies on manual feature selection approaches. Consequently, researchers have begun investigating the efficacy of very short-term recordings as well.
\citet{inoue2022development} used short-time frame ECG measurements from wearable devices with structured yoga exercises to build a thresholding Z-score-based classification model from Linear Discriminant Analysis (LDA), which achieved high screening accuracy within minutes of guided activity.
Recently, \citet{buza2023simple} used convolutional nearest neighbor for detecting psychiatric disorders using RRI time series recorded from the wearable Polar H10 device.
Likewise, \citet{ksikazek2025deep} explored a wearable short-term HRV assessment for psychiatric disorders, which groups schizophrenia and bipolar disorder together due to their shared ANS dysregulation. In their study of 60 individuals with the HRV-ACC dataset \cite{ksikazek10hrv}, ML models (including an ensemble of SVMs and a GRU-based neural network (NN)) trained on just 1-hour RRI segments could classify psychiatric disorders versus health controls with 80-83\% accuracy. 

Overall, these advances illustrate that wearable ECG/HRV devices, from adhesive chest patches, chest straps, and smartwatches, to everyday smart sensors, can capture clinically relevant HRV alterations in psychiatric disorders. The evolving landscape suggests promising opportunities for non-invasive, continuous monitoring outside traditional clinical settings. However, the challenge ahead lies in optimizing algorithms to maintain high accuracy with minimal data requirements and still be able to validate and contest the faithfulness and reliability of ``black-box'' models for clinical acceptance across larger and diverse patient populations in real-life settings.

\begin{table}[t!]
    \centering
    \caption{Summary of wearable ECG-based psychiatric disorder Detection Approaches}
    \renewcommand{\arraystretch}{1.2}
    \label{tab:my_label}
    \begin{tabularx}{\linewidth}{m{1.8cm}|m{1.5cm}|m{3cm}|m{2cm}|m{3cm}|m{2.5cm}}
    \hline
        \textbf{Author} & \textbf{Targets} & \textbf{Subjects\textsuperscript{[Dataset]}} & \textbf{Features} & \textbf{Methods} & \textbf{Devices}   \\
    \hline
        \citet{valenza2016predicting} & BD & BD: 14 & HRV features & SVM & Smartex PSYCHE wearable ECG   \\
    \hline
        \citet{ainunhusna2020bipolar} & BD & HC: 14, BD: 18 & MHR, HRV features & SVM & - \\
    \hline
        \citet{tiryaki2021real} & DP & Subjects: 79\hyperlink{fn:1}{\textsuperscript{[a]}}  & 1-lead ECG & CNN & VivaLNK Continuous ECG Recorder \\ 
    \hline
        \citet{zang2022end} & DP & HC: 37, DP: 37 & 2-lead ECG & CNN & - \\
    \hline
        \citet{tasci2024novel} 
        &         &         &          & ANN &  \\
    \cline{1-1} \cline{5-5}
        \citet{khare2023ecgpsychnet} & BD, DP, SZ  & HC: 35, BP: 62, DP: 17, SZ: 119\hyperlink{fn:2}{\textsuperscript{[b]}}  &  12-lead ECG & Hybrid ensemble classifiers & Philips ECG TC20 \\
    \cline{1-1} \cline{5-5}
        \citet{telangore2025novel} &  &  &  & Wavelet scattering network \& Fine KNN classifier & \\
    \hline
        \citet{corponi2024does} & BD & Subjects: 67 & PPG features & Bayesian hierarchical model & Empatica E4 smartwatch \\
    \hline
        \citet{buza2023simple} & & & & Convolutional nearest neighbor & \\
    \cline{1-1} \cline{5-5}
        \citet{ksikazek2025deep}  & SZ/BD & HC: 30, SZ/BD: 30\hyperlink{fn:3}{\textsuperscript{[c]}} & RRI time series & Ensemble of SVMs and GRU-based NN & Polar H10  \\
    \cline{1-1} \cline{5-5}
        \citet{nguyen2025human} & & & & Time-series Convolutional Attention NN & \\
    \hline
    \end{tabularx}
\end{table}
\hypertarget{fn:1}{\footnotetext[1]{PhysioNet European ST-T Database: \url{https://physionet.org/content/edb/1.0.0/}}}
\hypertarget{fn:2}{\footnotetext[2]{Psychiatry ECG: \url{https://kaggle.com/datasets/buraktaci/psychiatry-ecg}}}
\hypertarget{fn:3}{\footnotetext[3]{HRV-ACC: \url{https://zenodo.org/records/8171266}}}

\subsection{Explainable AI and Human-centered Contestable AI Systems in Healthcare}
\begin{figure}[t!]
    \centering
    \includegraphics[width=\linewidth]{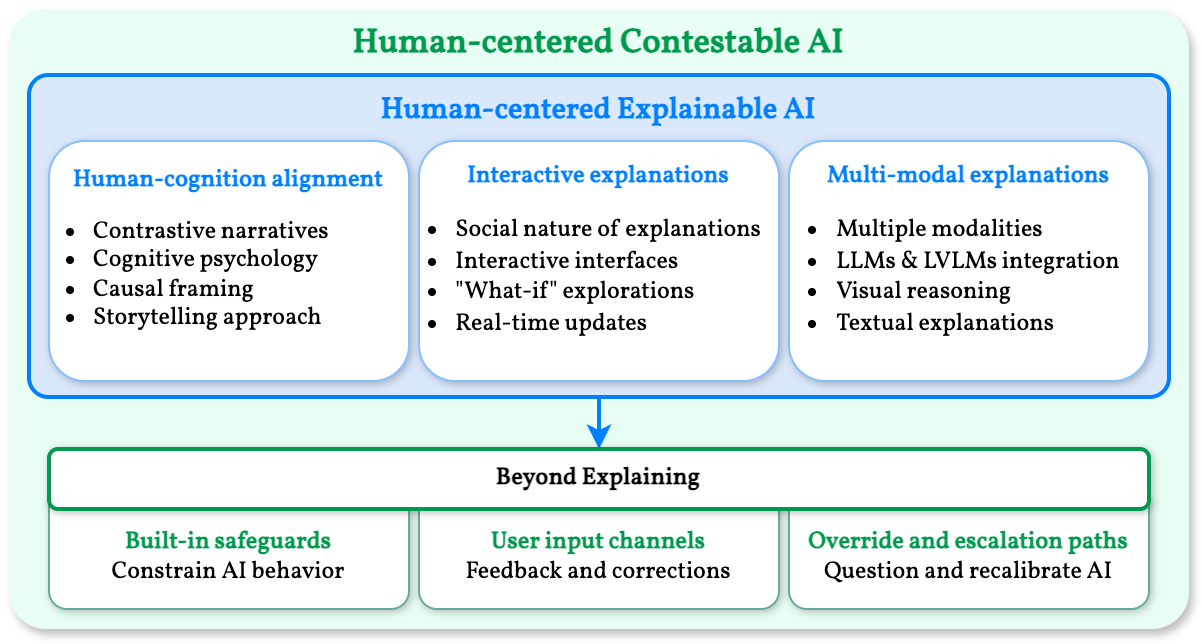}
    \caption{Evolution from Human-centered XAI toward Contestable AI Systems.}
    \label{fig:landscape}
\end{figure}

Despite several remarkable advances in ECG-based psychiatric disorder detection techniques, significant challenges remain before these technologies can be successfully translated into clinical practice.
The ``black-box'' nature of advanced AI techniques poses a barrier to clinical acceptance and implementation.
Hence, in the healthcare context, XAI has grown rapidly and proven particularly valuable, enhancing diagnostic accuracy, building trust among practitioners and patients, and ensuring ethical use of AI technologies \cite{nguyen2023towards,albahri2023systematic,vzlahtivc2023agile}. While explainability provides transparency, recent regulations and clinical realities demand that healthcare AI systems go further by empowering clinicians not only to understand AI decisions but also to meaningfully challenge and correct them, leading us to explore human-centered contestable AI systems, as illustrated in Figure~\ref{fig:landscape}.

\subsubsection{XAI Applications in Healthcare}
XAI applications aim to provide clinicians with meaningful insights into the decision-making process behind AI-driven diagnoses. 
Several families of XAI techniques have already been proposed and have proven their worth in clinical research. Post-hoc XAI methods (e.g., backpropagation-based \cite{simonyan2014deep,zeiler2014visualizing,binder2016layer}, Class Activation Mapping (CAM)-based \cite{selvaraju2020grad,nguyen2022secam,nguyen2024efficient,chattopadhay2018grad}, and perturbation-based \cite{truong2023towards,petsiuk2018rise,petsiuk2021black,lundberg2017unified,ribeiro2016should}) help clinicians verify \textit{where} and \textit{why} a black-box focuses. Prototype- and example-based methods ground predictions in comparators. While counterfactual explanations \cite{karimi2020model,mothilal2020dice} let clinicians explore actionable ``what-if'' scenarios. 
For psychiatric disorder prediction specifically, various XAI approaches have been applied across different data modalities.
\citet{arias2023enhancing} applied the SHapley Additive exPlanations (SHAP) \cite{lundberg2017unified} to XGBoost \cite{chen2016xgboost} for schizophrenia prediction based on key features in tabular format extracted from EEG signals.
\citet{jimenez2024machine} used Local Interpretable Model-Agnostic Explanations (LIME) \cite{ribeiro2016should} and SHAP on MRI scans for the schizophrenia classification. 
Misgar et al. \cite{misgar2024unveiling} applied GradCAM \cite{selvaraju2020grad} to visualize and understand a time series multi-branch DL's classification process on the motor activity data of people with psychiatric disorders.
Recent research demonstrates healthcare explainability advancing through four complementary approaches: (1) aligning with human cognition, (2) implementing interactive mechanisms for clinician engagement, and (3) utilizing multiple modalities for contextual justifications. These developments shift healthcare XAI from basic visualization techniques toward human-centered XAI systems integrated into clinical workflows, which are detailed as follows:
\begin{itemize}
    \item \textbf{Human-cognition and psychology alignment}: Modern XAI increasingly incorporates cognitive psychology principles to frame explanations that align with clinicians' mental models. For example, \citet{zhang2022towards} proposed structures contrastive ``why A not B'' narratives around the perceptual stages of attention, appraisal, and reasoning, to generate relatable explanations inspired by humans' perceptual process from cognitive psychology. Reviews of psychology-based AI indicate that empathy, familiarity, and causal framing are essential for human-aligned conversational explanations, particularly in emotionally charged contexts \cite{dazeley2021levels}. This approach emphasizes storytelling, causal reasoning, and domain-specific language rather than gradient-based visualizations.  
    \item \textbf{Interactive and human-in-the-loop explanations:} Interactive explanations are often seen in human-human interaction, which is a social nature of explanations \cite{miller2019explanation}. 
    Static explanations are evolving into interactive systems where users can interact with the AI's explanation through an interface, such as changing the attribute values of an instance \cite{cheng2019explaining} or creating theory-driven explanations \cite{wang2019designing}, to inspect the updated prediction. For example, the CLARUS platform enables free-form ``what-if'' counterfactual questions (e.g., lowering blood-pressure values) with immediate risk curve updates, improving correction accuracy by 27\% in diabetes research \cite{metsch2024clarus}. 
    \item \textbf{Multi-modal and modality-adaptive explanations:} Multi-modal explanations involve multiple modalities in the primary and the explanation tasks. For instance, several works blended multi-modal concept-based explanations with LLMs or large vision language models (LVLMs), which have emerged as powerful tools for combining language understanding with visual reasoning, (i.e., visual question answering and multi-modal learning \cite{nguyen2024langxai,nguyen2024xedgeai,patricio2025cbvlm,nguyen2025human}).
    Through a user study, \citet{robertson2021wait} found that when users' attentional resources were limited, presenting explanation interventions via different modalities, like audio, interactivity, and text, enhanced real-time comprehension.
    These capabilities present new opportunities for enhancing explainability in visual perception tasks through textual explanations, advancing human-centered XAI systems \cite{nguyen2024langxai,nguyen2024xedgeai}.
\end{itemize}

\subsubsection{Beyond Explaining: Towards Human-Centered Contestable AI Design in Healthcare}\label{sec:conAIrw}
\textit{\textbf{Contestability}} goes one step further than explainability: beyond merely understanding the AI, human users must be able to question, intervene in, and correct the AI's decisions. 

Recent regulations like the GDPR \cite{regulation2016regulation} and the EU AI Act \cite{neuwirth2022eu} emphasize the importance of model interpretability in healthcare applications. A requirement of transparency or explainability can be found in legislation. Articles 13 and 14 of the GDPR specify that if data subjects are profiled, they have a right to ``meaningful information about the logic involved.'' This applies to the medical context as explained in the official EU guideline.
GDPR's requirement for explainability should be understood as a requirement for contestability, where AI decision-making must be explainable to a degree that makes it possible for an individual to contest the decision of the system. 
Specifically, Article 22 states that in cases where a data subject may legitimately be subjected to automated decision-making, including profiling, the data controller should safeguard the data subject's right ``to express his or her point of view and to \textit{contest} the decision'' \cite{regulation2016regulation}.
Similarly, the Canadian Directive on Automated Decision-Making \cite{board_board_2019}, contestability is directly addressed through specific requirements related to the recourse mechanism, where Section 6.4 explicitly states ``providing clients with any applicable recourse options that are available to them to \textit{challenge} the administrative decision.'' 
Other regulations from such as Health Canada \cite{canada2025}, UK's MHRA~\cite{and_2024}, U.S FDA \cite{center_2025}, or Montréal Declaration on Responsible AI \cite{de_2023} highlight facets of contestability, from rights to human oversight and explanation to mechanisms for appeal, in the context of AI and automated decision-making.
As a response to recent regulations that emphasize the ability to contest the AI decision and make explanations to be more human-oriented \cite{miller2019explanation}, contestable AI goes beyond XAI by enabling users to actively challenge and dispute those decisions based on the provided explanations, essentially giving them the ability to contest the AI's output in a meaningful way. It is not just about understanding the reasoning but also having the agency to question and potentially overturn a decision if necessary \cite{lyons2021conceptualising,alfrink2023contestable}.
Based on the regulatory frameworks presented that consistently emphasize human agency in questioning AI decisions, from GDPR's \textit{``right to contest,''} to Canada's requirement for \textit{``challenge''} mechanisms, to various health authorities' emphasis on oversight and appeals, we propose the following definition of a contestable AI System:

\begin{tcolorbox}[colback=blue!5!white,colframe=blue!55!black,title=\textbf{Definition}]
  \textbf{\textit{Contestable AI System:}} An interactive computational system that allows human challenge throughout its decision-making lifecycle, maintains conscious transparency of its operational processes, and incorporates a safeguard mechanism to constrain algorithmic behavior.
\end{tcolorbox}

In healthcare AI systems, a contestable AI system invites a form of dialogue or challenge, wherein clinicians, or even patients, can disagree with an algorithm's output and have that dissent meaningfully influence the outcome. 
In clinical practice, this concept aligns with the reality that diagnoses and treatments are often reached collaboratively or iteratively, where the second options are sought, and the initial ones are refined. 
Similarly, an AI diagnostic recommendation should not be seen as final. 
Concrete examples of contestable AI in healthcare are still emerging, but some prototypes point to effective strategies.
\citet{hirsch2017designing} described an automated psychotherapy feedback system wherein the system's evaluations of therapy sessions are made legible and open to dispute by therapists. Their design included providing detailed explanations for each assessed metric and even allowing users to trace the AI's reasoning down to raw transcript data. Crucially, the system offered a mechanism for the human user to record disagreements or add context that the AI might have missed. 
This kind of interactive contestability ensures that the human expert remains in control, where the AI assists with analysis, but the therapist can correct the record, preventing blind reliance on an imperfect algorithm.
\citet{ploug2020four} argued for a patient-centric approach in contestable AI diagnostics, identifying dimensions such as informing users about the system’s data usage, biases, accuracy, clinical validity, and clearly outlining the AI’s role as prerequisites to enable effective contestation.
In essence, a mental health AI tool should communicate its limitations (e.g., trained primarily on young- and middle-aged, rendering them less reliable for older adult populations) so that clinicians know when to question it, and it should be embedded in a care workflow that allows those questions to alter decisions (e.g., a questionable algorithmic output triggers a thorough case review by a human clinician before any action).

In summary, making AI systems explainable and contestable in the healthcare context, especially in mental health, is now recognized as vital for safe, ethical, and effective deployment.
Explainability provides transparency and insight, helping clinicians integrate AI recommendations into their decision-making process with appropriate confidence. Contestability ensures that clinicians and patients remain empowered, so that the AI is a tool at their service rather than an opaque authority. Together, a human-centered contestable AI system should emphasize features such as: 
\begin{enumerate}
    \item \textbf{Human-centered explanations} to ground any contestation, because clinicians can only know when to contest a decision if they understand the rationale behind models, even without or with very little knowledge about AI and XAI.
    \item \textbf{Built-in safeguards} to constrain the behavior of AI systems, where procedural safeguards restrict unilateral AI systems' decisions. One type of safeguard involves creating a self-adversarial decision-making process \cite{alfrink2023contestable}. Another can be achieved by introducing a second automated system external to the controlling organization, through which machine decisions are made. When inconsistencies arise between the primary system and verification mechanisms, the case can either be escalated for human expert evaluation or handled through built-in conflict resolution protocols.
    \item \textbf{User input channels} for feedback and corrections (e.g., a clinician can provide an alternative suggestion for AI decisions, and the AI system should acknowledge this input by updating the model with new information or logging the disagreement for review).
    \item \textbf{Override and escalation paths} to ask follow-up questions to the model's explanation, the option to validate, adjust or correct input data and see how the output changes, or a ``reject/override'' button that either recalibrates the AI’s recommendation or routes the decision to a human supervisor.
\end{enumerate}

%% file: sec/3_method.tex
\section{Methodology}\label{sec:method}
This section presents the design of Heart2Mind, our human-centered contestable psychiatric disorder diagnosis system that leverages wearable ECG monitors for objective assessment. We first provide an overview of the complete framework architecture, detailing how cardiac monitoring, AI-based detection, and contestable diagnosis components work together to create a clinically viable system. We then focus on our novel proposed principle and design of the contestable LLMs system, which enables healthcare professionals to actively challenge and validate AI decisions through natural language interaction. By combining self-adversarial explanations with interactive contestation mechanisms, our methodology addresses the critical need for transparency and human oversight in AI-assisted mental health diagnostics.

\subsection{Overview of Heart2Mind Framework}
\begin{figure}[ht!]
    \centering
    \includegraphics[width=\linewidth]{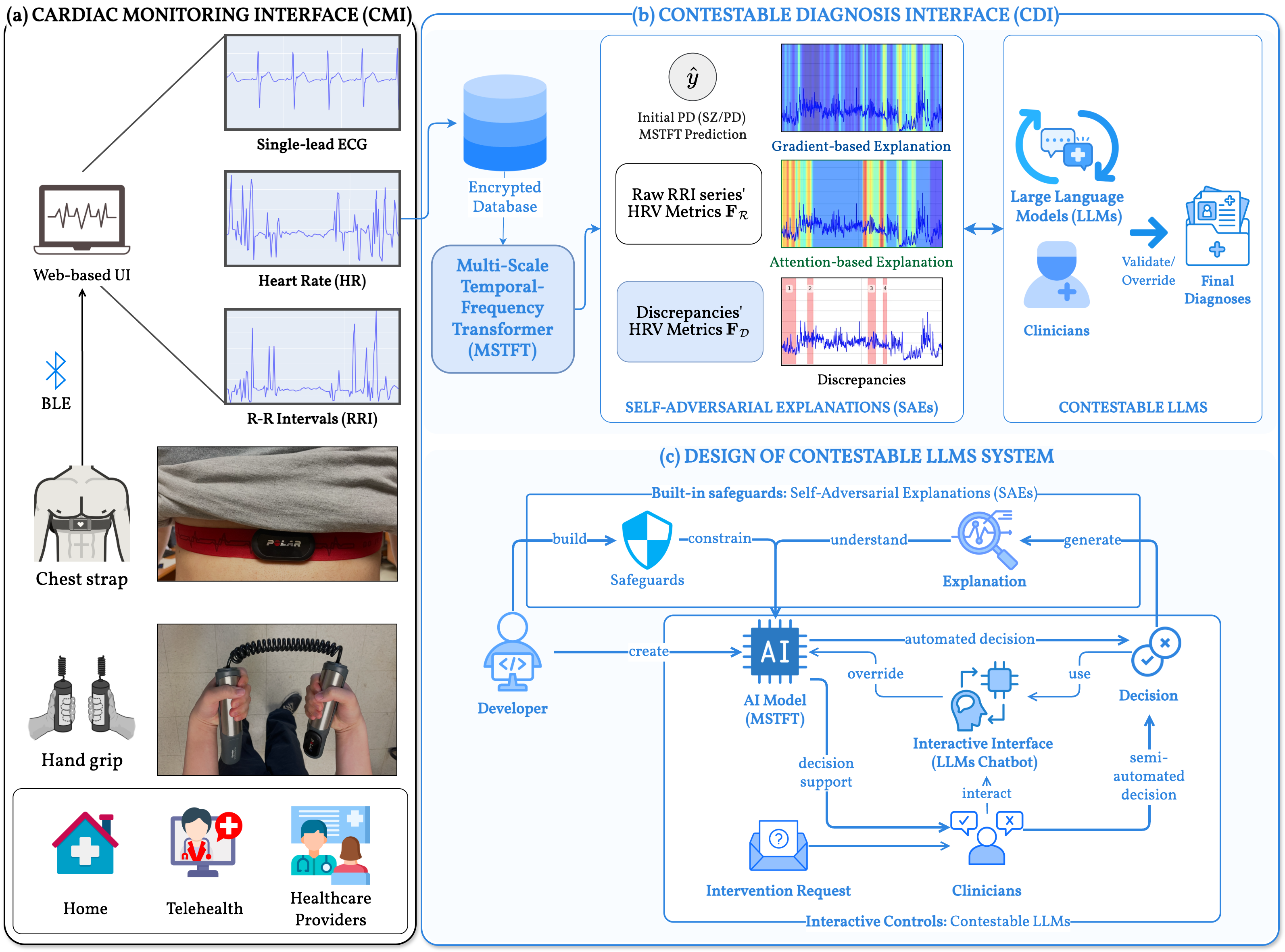}
    \caption{The overview of Heart2Mind framework: (a) Cardiac Monitoring Interface (CMI), (b) Contestable Diagnosis Interface (CDI), and (c) the design of contestable LLMs system.}
    \label{fig:overview}
\end{figure}

Figure~\ref{fig:overview} provides an overview of our proposed Heart2Mind framework, which integrates two core interfaces to create a comprehensive human-centered contestable psychiatric disorder diagnosis system.

\begin{enumerate}
    \item \textbf{Cardiac Monitoring Interface (CMI) (Figure~\ref{fig:overview}a):} This web-based interface allows diverse user types to capture real-time cardiac signals (ECG/HR/RRI) from wearable devices such as the Polar H9/H10 chest strap and hand grip sensors. The interface seamlessly connects with these devices via Bluetooth Low Energy (BLE), processes the raw signals, and stores the data in an encrypted database. Sessions can be conducted in clinical settings under the direct supervision of healthcare providers, in remote telehealth scenarios with virtual professional oversight, or independently by users in their home environments. 

    \item \textbf{Contestable Diagnosis Interface (CDI) (Figure~\ref{fig:overview}b):} This web-based interface provides clinicians the MSTFT diagnosis for psychiatric disorders with the explainability and contestability for the AI's decisions. Specifically designed for mental health professionals, this interface enables these practitioners to interpret, evaluate, or override AI's diagnoses:
    \begin{enumerate}
        \item \textbf{Multi-Scale Temporal-Frequency Transformer (MSTFT)}: At the core of our diagnosis system is the MSTFT, which processes the raw RRI time series to classify them as ``control'' (healthy) or ``treatment'' (schizophrenia/bipolar disorder). The model integrates multi-scale temporal features and frequency-domain wavelet features through a transformer-based architecture, utilizing cross-attention fusion and self-attention mechanisms to achieve superior classification performance.
        \item \textbf{Self-Adversarial Explanations (SAEs):} SAEs generate dual visual explanations from attention-based and gradient-based explanation methods. Then, SAEs identify regions of discrepancy by comparing two visual explanations where the model's decision-making process shows internal inconsistencies, flagging potentially unreliable predictions for clinicians' review.
        \item \textbf{Contestable LLMs:}  A chatbot interface powered by LLMs allows clinicians to interact with and challenge the MSTFT's decisions. The LLM analyzes the model prediction, HRV metrics, and identifies discrepancies to provide natural language explanations and enable clinicians to validate or overturn initial predictions.
    \end{enumerate}

\end{enumerate}

\subsection{Design of Contestable LLMs System}
While multiple dimensions of contestation exist (as introduced in Section~\ref{sec:conAIrw}), our work focuses specifically on enabling stakeholders to challenge performance and validate AI decisions through contestable LLMs as a chatbot interface.
We introduce the contestable LLMs system, based on integrating LLMs with contestable AI principles, creating an intelligent conversational interface that empowers clinicians to actively challenge and examine AI-driven psychiatric diagnoses. This innovative approach transforms traditional explainable AI systems into truly interactive decision-support tools. The Contestable LLMs analyze the MSTFT model's predictions, comprehensive HRV metrics, and regions of explanation discrepancy identified by SAEs to generate contextualized natural language explanations. Unlike conventional AI explanations that merely present information, our contestable LLMs engage in dynamic dialogue with clinicians, allowing them to question specific aspects of the diagnosis, request additional clarifications, and ultimately validate or overturn the initial AI prediction. This bidirectional interaction creates a collaborative decision-making process where clinical expertise and AI capabilities synergistically converge to enhance diagnostic accuracy and maintain human oversight in critical healthcare decisions.

As illustrated in Figure \ref{fig:overview}c, our contestable LLMs system implements two core mechanisms for high contestability, building on \citet{alfrink2023contestable} foundational work on contestable AI design as follows:
\begin{enumerate}
    \item \textbf{Built-in safeguards with SAEs:} Developers implement built-in safeguard mechanisms that prevent unilateral decision-making by the AI. Our preferred approach involves creating a self-adversarial decision-making mechanism to validate AI decisions via explanation maps generated by XAI methods. If any unreliability or lack of faithfulness in the AI's decisions is detected, these cases are flagged for human expert review, or built-in conflict resolution systems are activated to address the discrepancy.
    \item \textbf{Interactive Controls with Contestable LLMs:} Human controllers use a chatbot as the interactive interface to correct or override AI system decisions. Clinicians use interactive controls, explanations, and intervention requests to contest AI system decisions. Explanations should contain the information necessary for a decision, subject to the exercise of their rights to human intervention and contestation. The final decision would be the result of a negotiation between the system and clinicians.
\end{enumerate}

%% file: sec/4_imp.tex
\section{Implementation}\label{sec:impl}
Having established the conceptual framework and design principles of Heart2Mind in the previous section, we now detail the technical implementation of each component in our human-centered contestable psychiatric disorder diagnosis system. Our implementation consists of three primary components: First, the Cardiac Monitoring Interface (CMI) provides the data acquisition layer, capturing cardiac signals from wearable devices through a secure and user-friendly interface. Second, the Multi-Scale Temporal-Frequency Transformer (MSTFT) serves as the core model, processing raw physiological signals to provide psychiatric disorder predictions. Finally, the Contestable Diagnosis Interface (CDI) creates a human-AI interaction layer, implementing our novel Self-Adversarial Explanations (SAEs) and Contestable LLMs to ensure transparency, interpretability, and contestability.

\begin{figure}[b!]
    \centering
    \includegraphics[width=\linewidth]{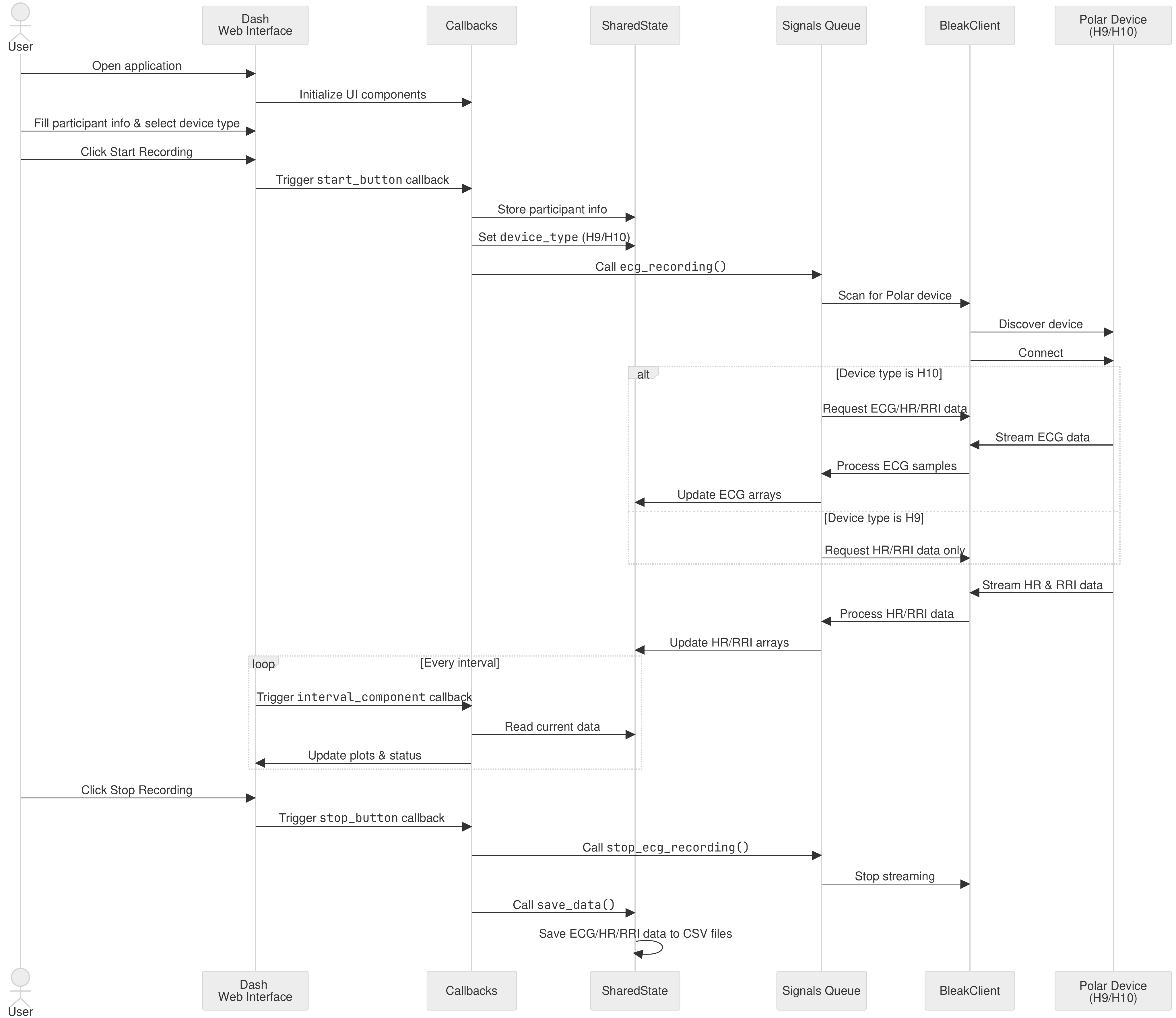}
    \caption{Sequence diagram illustrating the cardiac signal recording workflow in the Cardiac Monitoring Interface (CMI). The diagram shows the interaction flow between user actions, web interface components, and Polar device communication, including device-specific branching for H9 (HR/RRI only) and H10 (ECG/HR/RRI) configurations.}
    \label{fig:cmi_flow}
\end{figure}

\subsection{Cardiac Monitoring Interface (CMI)}
The implementation of our Heart2Mind framework begins with the CMI, which serves as the foundational interface for capturing high-quality cardiac data essential for psychiatric disorder detection. The CMI is designed to seamlessly integrate wearable technologies with data acquisition and processing functionalities, ensuring real-time and continuous monitoring of cardiac signals (i.e., ECG/HR/RRI).

\begin{figure}[t!]
    \centering
    \includegraphics[width=\linewidth]{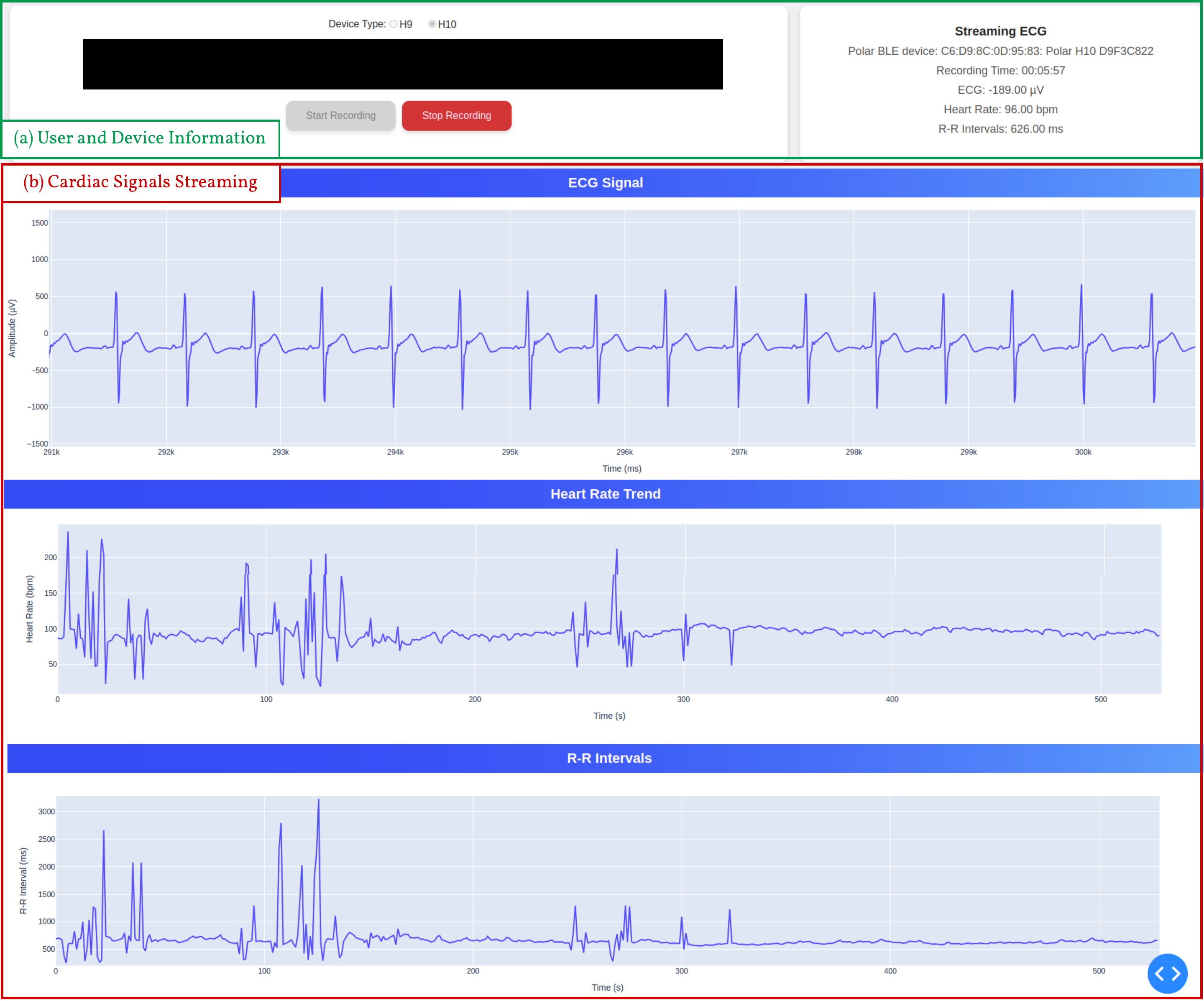}
    \caption{The dashboard of the Cardiac Monitoring Interface (CMI) showing two main panels: (a) User controls and device status indicators for clinical monitoring, (b) Cardiac signals displaying ECG signal, HR, and RRI time series.}
    \label{fig:cmi_ui}
\end{figure}

\subsubsection{Wearable ECG Monitor}
To demonstrate our monitoring system, we employ the Polar H9 or Polar H10 sensors as the standard cardiac monitoring devices, selected for their proven high precision in single-lead ECG signal acquisition and heart rate variability measurements \cite{li2023evaluation,speer2020measuring}. 
Both devices capture HR in beats per minute (BPM) and RRI in milliseconds (ms) with a 1-second sampling rate.
Polar H10 offers additional capabilities above Polar H9, recording single-lead ECG at 130 Hz with measurements in microvolts ($\mu V$). providing richer physiological data for the analysis.

\subsubsection{Interface}
Building upon the BleakHeart library~\cite{smeraldi2025}, we developed a monitoring interface that seamlessly integrates with wearable devices to capture real-time cardiac signals and securely store them in an encrypted database. 
Figure~\ref{fig:cmi_flow} illustrates the detailed flow of a recording session, showcasing our modular architecture that separates data acquisition, state management, and user interface components.

The recording process begins with users self-entering their personal information, including their name, age, and sex, which is immediately encrypted using a unique ID key to ensure privacy compliance. Users then select their device type (H9 or H10) to initiate the recording session.
The CMI automatically scans for available Polar devices via BLE and establishes a wireless connection. 
Once connected, the system configures appropriate data streams based on the device capabilities (ECG/HR/RRI for the H10 and HR/RRI only for the H9).

During the session, users can follow light free-living protocols that include short corridor walks interleaved with seated rest periods where they can sit down doing daily business tasks.
Raw sensor data undergoes real-time processing with timestamps synchronized to the host system clock, ensuring temporal accuracy crucial for HRV analysis. 
The processed cardiac signals are simultaneously streamed to the web-based user interface, as shown in Figure~\ref{fig:cmi_ui}, which displays ECG waveforms, heart rate, and RRI time series in an intuitive format for clinical observation.
Upon session completion, typically after a predetermined duration (at least 70 minutes), the interface prompts users to stop recording. The collected cardiac signals are then automatically saved in a structured CSV format and securely transmitted to an encrypted database accessible to authorized clinicians.
This comprehensive data capture and storage approach ensures both the integrity of the physiological measurements and compliance with healthcare data protection standards.

\subsection{Multi-Scale Temporal-Frequency Transformer (MSTFT)}

\begin{figure}
    \centering
    \includegraphics[width=\linewidth]{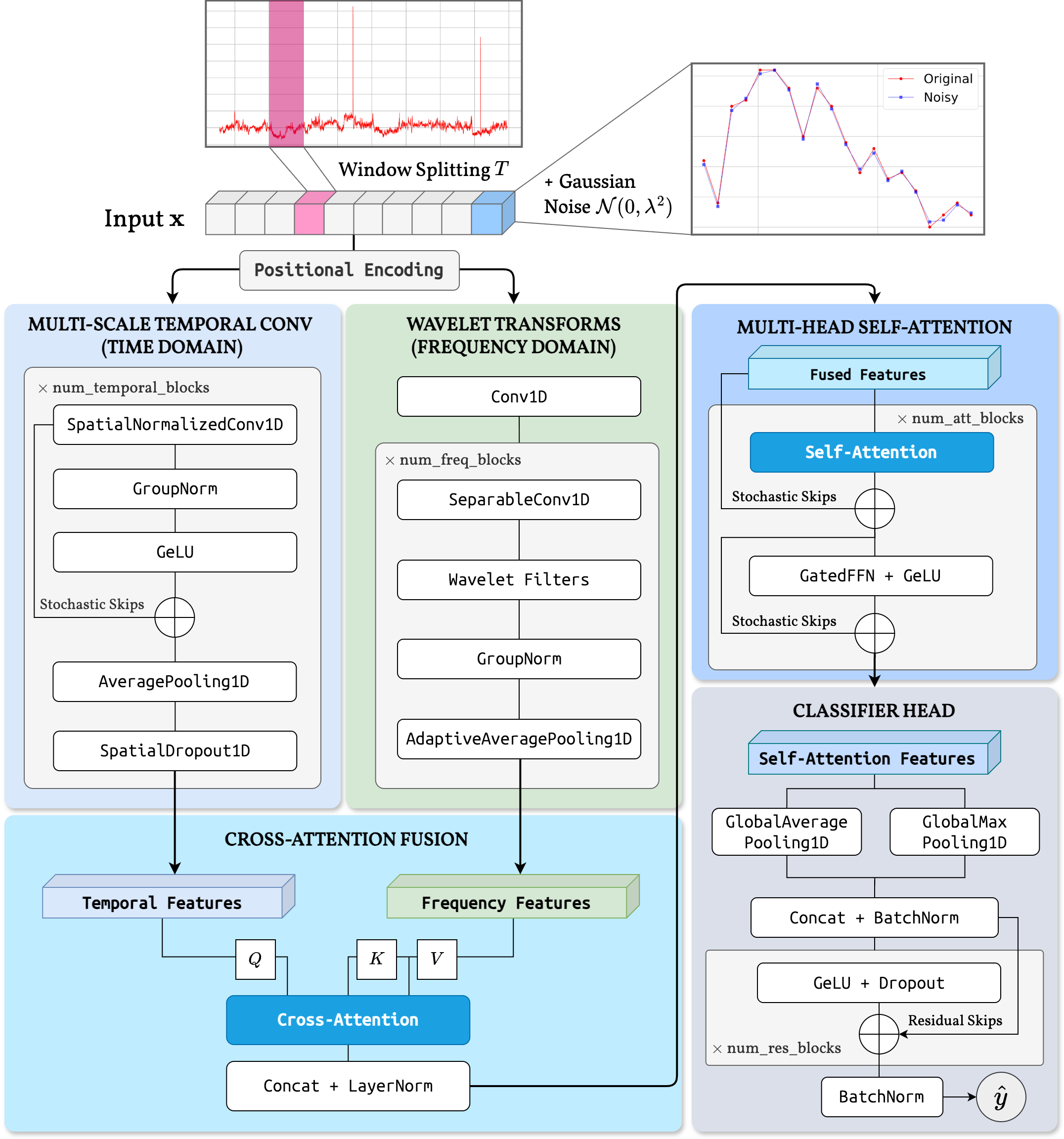}
    \caption{Architecture overview of the Multi-Scale Temporal-Frequency Transformer (MSTFT). The model processes raw RRI time series through parallel temporal and frequency pathways, integrating features via cross-attention fusion before final classification. Key components include noise-augmented input preprocessing, multi-scale temporal convolutions with stochastic skips, wavelet-based frequency transforms, and attention-based feature fusion blocks.}
    \label{fig:mstft_model_desc}
\end{figure}
Building upon the cardiac data captured by the CMI, we propose the MSTFT, the core diagnosis model of our Heart2Mind framework. As illustrated in Figure~\ref{fig:mstft_model_desc}, MSTFT integrates multi-scale temporal features and frequency-domain wavelet features through a transformer-based architecture (i.e., cross-attention fusion and self-attention mechanism) to classify RRI time series as either healthy ``control'' or ``treatment''.
Our architecture comprises several blocks, each designed to extract temporal and spectral patterns within input sequences.

\subsubsection{Input Preprocessing and Positional Encoding}
Our model begins by processing the raw RRI time series input $\mathbf{x} = (\mathbf{x}_1, \mathbf{x}_2, \dots, \mathbf{x}_{T}) \in \mathbb{R}^{T \times 1}$ of length $T$.
Recognizing cardio signals recorded from wearable devices often contain noise artifacts, we enhance model robustness by introducing controlled Gaussian noise $\mathcal{N}(0, \lambda^2)$ to the input sequence $\mathbf{x}$, providing regularization against noise sensitivity during the training phase.
Then, the positional encoding $\mathbf{P} \in \mathbb{R}^{T \times d}$ projects the input to higher-dimension space as follows:
\begin{gather}
\tilde{\mathbf{x}} = \mathcal{C}(\mathbf{x} + \epsilon) + \mathbf{P}, \quad \epsilon \sim \mathcal{N}(0, \lambda^2), \\
\mathbf{P}_{(t,2k)} = \sin\left(\frac{t}{\tau^{2k/d}}\right), \quad  \mathbf{P}_{(t,2k+1)} = \cos\left(\frac{t}{\tau^{2k/d}}\right),
\end{gather}
where $\tau$ controls wavelength progression and $\mathcal{C}$ denotes the embedding 1D convolutional layer that transforms the input to dimension $d$.

\subsubsection{Multi-Scale Temporal Convolutions Block (Time Domain)}
The temporal block captures cardiac rhythm patterns across multiple scales through dilated convolutions.
Starting with the encoded input $\mathbf{Z}_t^{(0)}=\tilde{\mathbf{x}}$, each layer applies convolutions with exponentially increasing dilation rates:
\begin{equation}
\mathbf{Z}_t^{(i+1)} = S\left(\mathbf{Z}_t^{(i)}, \text{GrpNorm}\left(\sigma_g\left(\mathcal{C}_\mathcal{D}^{(d=2^i)}\left(\mathbf{Z}_t^{(i)}\right)\right)\right)\right), \quad \forall i \in \{0, \dots, n_t-1\},
\end{equation}
where $\mathcal{C}_\mathcal{D}^{(d=2^i)}$ represents causal dilated convolutions with dilation rate $2^i$, $\text{GrpNorm}(\cdot)$ provides group layer normalization, $\sigma_g(\cdot)$ is the GELU activation function, $n_t$ is the total number of temporal blocks, and $S$ is the stochastic skips learning block. Each subsequent block reduces the number of filters by half to ensure multi-scale representation learning. Finally, the temporal features $\mathbf{Z}_t$ are achieved.

\paragraph{Stochastic Skips Learning}
\begin{figure}
    \centering
    \includegraphics[width=0.8\linewidth]{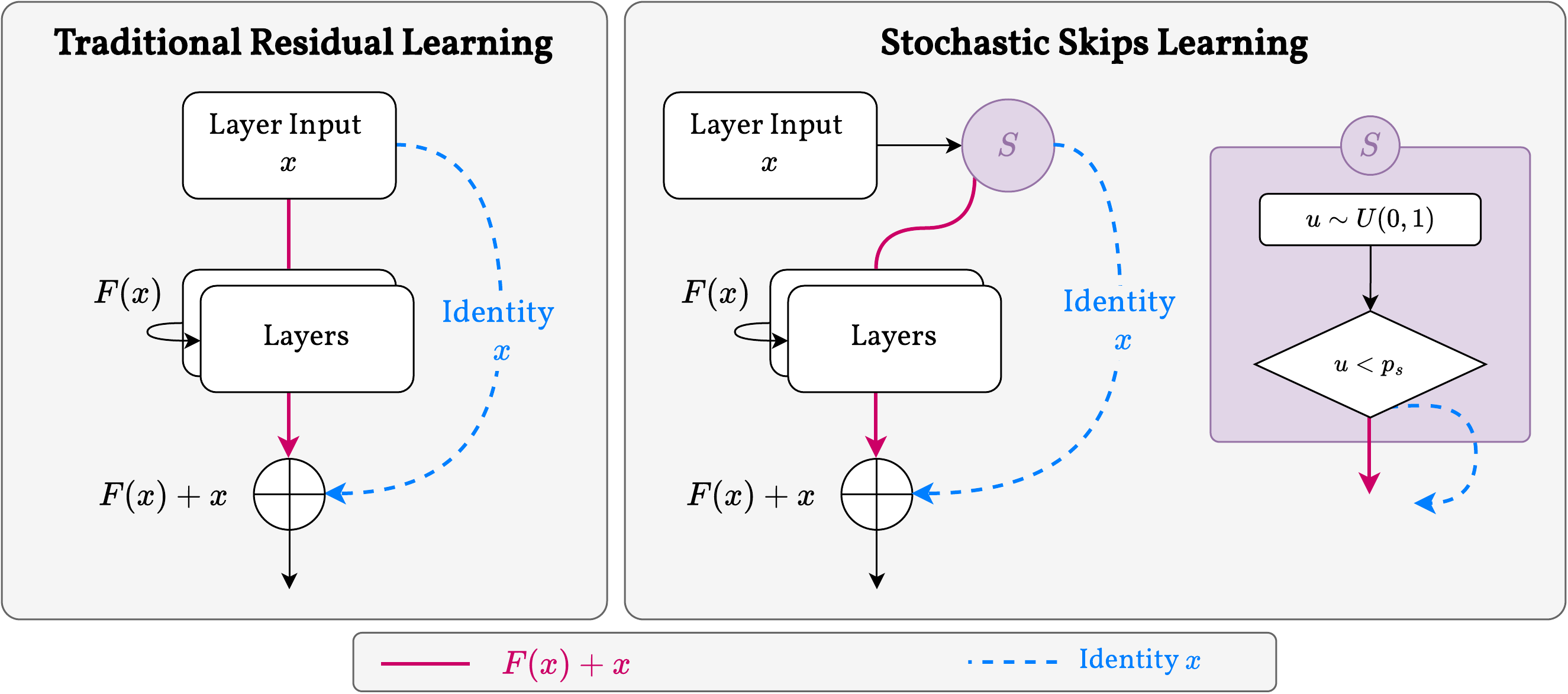}
    \caption{Comparison of traditional residual learning with stochastic skips learning mechanism. While traditional residual connections deterministically add the transformed features $F(x)$ to the input $x$, stochastic skips learning introduces a probabilistic decision gate controlled by survival probability $p_s$, randomly bypassing transformations during training to prevent overfitting while maintaining gradient flow.}
    \label{fig:stskips}
\end{figure}
Unlike traditional residual connections that deterministically add the input to the transformed output, our stochastic skips module introduces probabilistic regularization (Figure \ref{fig:stskips}):
\begin{equation}
S(x, F(x)) =
\begin{cases}
F(x)+x, & \text{if } u < p_s ,\quad u\sim U(0,1)\\
x, & \text{otherwise}
\end{cases}
\end{equation}
where $p_s$ represents the survival probability of the stochastic residual path, $u\sim U(0,1)$ denotes a random sample from a uniform distribution between 0 and 1. This stochastic depth mechanism acts as implicit regularization by randomly dropping residual connections during training, thus effectively preventing overfitting while maintaining gradient flow.

\subsubsection{Wavelet Transforms Block (Frequency Domain)}
In addition to the temporal analysis, the frequency domain block extracts spectral features via learnable wavelet transforms modeled as separable convolutions:
\begin{gather}
\mathbf{Z}_f^{(0)} = \mathcal{C}_f(\tilde{\mathbf{x}})\\ \mathbf{Z}_f^{(j+1)} = \text{GrpNorm}\left(\sigma_g\left(\mathcal{C}_\mathcal{S}\left(\mathbf{Z}_f^{(j)}\right)\right)\right), \quad j \in \{0, \dots, n_f-1\},
\end{gather}
where $\mathcal{C}_f$ initializes frequency-specific 1D convolutional embeddings, $\mathcal{C}_{S}$ is the 1D separable convolutional layer, $n_f$ is the total frequency blocks. Adaptive average pooling adjusts frequency-domain features to match temporal feature dimensionality to achieve the frequency features $\mathbf{Z}_f$.

\subsubsection{Cross-Attention Fusion Block}
The cross-attention fusion block serves as a critical component for integrating information from the temporal and frequency domains. 
This block employs an efficient cross-attention mechanism that enables linear-time computation complexity rather than quadratic complexity \cite{chen2021crossvit}, and allows for the effective integration of complementary information from both domains while preserving the original feature representations \cite{li2024crossfuse}.
Given the temporal features $\mathbf{Z}_t$ and frequency features $\mathbf{Z}_f$, we first project them to a common embedding space with dimension $d_p$ using linear transformations $\mathbf{H}_t = \mathcal{W}_t\mathbf{Z}_t, \mathbf{H}_f = \mathcal{W}_f\mathbf{Z}_f$, where $\mathcal{W}_t$ and $\mathcal{W}_f$ are learnable projection matrices. The cross-attention mechanism then operates by treating the temporal features as queries and the frequency features as keys and values:
\begin{equation}
\mathbf{Q} = \mathbf{H}_t\mathbf{W}^Q, \quad \mathbf{K} = \mathbf{H}_f\mathbf{W}^K, \quad \mathbf{V} = \mathbf{H}_f\mathbf{W}^V
\end{equation}
where $\mathbf{W}^Q$, $\mathbf{W}^K$, and $\mathbf{W}^V$ are learnable weight matrices for query, key, and value projections, respectively. The cross-attention operation is then computed as:
\begin{equation}
\text{CrossAttn}(\mathbf{Q}, \mathbf{K}, \mathbf{V}) = \text{softmax}\left(\frac{\mathbf{Q}\mathbf{K}^T}{\sqrt{d_k}}\right)\mathbf{V}
\label{eq:crossattn}
\end{equation}
where $d_k$ is the dimension of the key vectors. To capture more complex relationships, we employ multi-head attention:
\begin{equation}
\mathbf{F}_{\text{multi}} = \text{Concat}(\text{h}_1, \ldots, \text{h}_h)\mathbf{W}^O, \quad \text{h}_i = \text{CrossAttn}(\mathbf{Q}\mathbf{W}_i^Q, \mathbf{K}\mathbf{W}_i^K, \mathbf{V}\mathbf{W}_i^V)
\end{equation}

The fused representation $\mathbf{F}$ is then obtained by concatenating the multi-head attention output with the original temporal and frequency features, followed by layer normalization:
\begin{equation}
\mathbf{F} = \text{LayerNorm}(\text{Concat}[\mathbf{F}_{\text{multi}}, \mathbf{H}_t, \mathbf{H}_f])
\end{equation}

\subsubsection{Multi-Head Self-Attention Block}
Multi-Head Self-Attention Block enhances the fused representations by capturing long-range dependencies within the integrated features. 
Unlike standard transformer blocks, our implementation incorporates specialized components to improve performance for RRI sequence classification.

Given the fused representation $\mathbf{F}$ from the Cross-Attention Fusion Block, the multi-head self-attention operation is applied as follows:
\begin{equation}
\mathbf{A} = \text{SelfAttn}(\mathbf{F}, \mathbf{F}, \mathbf{F})
\label{eq:multihead}
\end{equation}
where the multi-head self-attention is computed similarly to the cross-attention mechanism, but with the same sequence serving as queries, keys, and values. To enhance the expressiveness of the model, we incorporate an attention-gating mechanism $\mathbf{G} = \sigma(\mathbf{W}_g\mathbf{A} + \mathbf{b}_g)$, where $\mathbf{W}_g$ and $\mathbf{b}_g$ are learnable parameters, and $\sigma$ is the sigmoid activation function. The gated attention output is then computed as $\mathbf{A}_g = \mathbf{A} \odot \mathbf{G}$ where $\odot$ denotes element-wise multiplication. To maintain stability during training, we employ channel-wise scaling with a learnable parameter $\alpha$ initialized to a small value $\mathbf{F}' = \mathbf{F} + \alpha \mathbf{A}_g$.

Following the scaled addition, we apply a position-wise feed-forward network with an expansion factor to increase the model's capacity. The final output of the self-attention block is:
\begin{equation}
\mathbf{F}'' = \text{LayerNorm}(\mathbf{F}' + \text{Dropout}(\text{FFN}(\mathbf{F}')))
\end{equation}

This multi-head self-attention mechanism effectively captures complex temporal-frequency dependencies in the fused representation, allowing the model to learn intricate patterns in RRI time series that are crucial for accurate classification.
\subsubsection{Classifier Head Block}
Classifier Head Block transforms the representation learned by the Multi-Head Self-Attention Block into the final classification decision. This block employs multiple techniques to maximize information extraction and classification performance.

First, we apply both global average pooling and global max pooling operations to capture different aspects of the sequence representation. These pooled representations are concatenated to form a comprehensive feature vector.
These pooled representations are concatenated to form a comprehensive feature vector.
To ensure stable training, we apply batch normalization to the concatenated features.

\begin{gather}
\mathbf{h}_{\text{avg}} = \frac{1}{T}\sum{t=1}^{T}\mathbf{F}''t, \quad \mathbf{h}_{\text{max}} = \max_{t \in {1,\ldots,T}}\mathbf{F}''_t \\
\mathbf{h} = \text{Concat}[\mathbf{h}_{\text{avg}}, \mathbf{h}_{\text{max}}]\\
\mathbf{h}_{\text{norm}} = \text{BatchNorm}(\mathbf{h})
\end{gather}

The normalized features then pass through a series of dense layers with attention-weighted residual connections. This structure enhances feature extraction while maintaining gradient flow during training:
\begin{gather}
\mathbf{h}_1 = \sigma_g(\mathbf{W}_1\mathbf{h}_{\text{norm}} + \mathbf{b}_1)\\
\mathbf{a} = \sigma(\mathbf{W}_a\mathbf{h}_1 + \mathbf{b}_a)\\
\mathbf{h}_{\text{att}} = \mathbf{h}_1 \odot \mathbf{a}\\
\mathbf{h}_{\text{res}} = \mathbf{W}_r\mathbf{h}_{\text{norm}} + \mathbf{b}_r\\
\mathbf{h}_{\text{out}} = \text{GrpNorm}(\mathbf{h}_{\text{att}} + \mathbf{h}_{\text{res}})
\end{gather}
where $\sigma_g$ is the GELU activation function, and $\odot$ denotes element-wise multiplication. The attention mechanism allows the model to focus on the most discriminative features for classification.

Finally, the output layer produces the probability of the input RRI sequence belonging to the positive class:
\begin{equation}
\hat{y} = \sigma(\mathbf{w}_o^T\mathbf{h}_{\text{out}} + b_o)
\end{equation}

\subsection{Contestable Diagnosis Interface (CDI)}
In this section, we illustrate our detailed implementation of the Contestable Diagnosis Interface (CDI), providing clinicians with a comprehensive platform to monitor the psychiatric disorder detection results of MSTFT, observe the explanations and discrepancies with SAEs, and challenge AI-driven diagnoses with the contestable LLMs system.
This holistic design ensures that clinical expertise remains central to the diagnostic process while leveraging the capabilities of AI.

\begin{figure}[ht!]
    \centering
    \includegraphics[width=\linewidth]{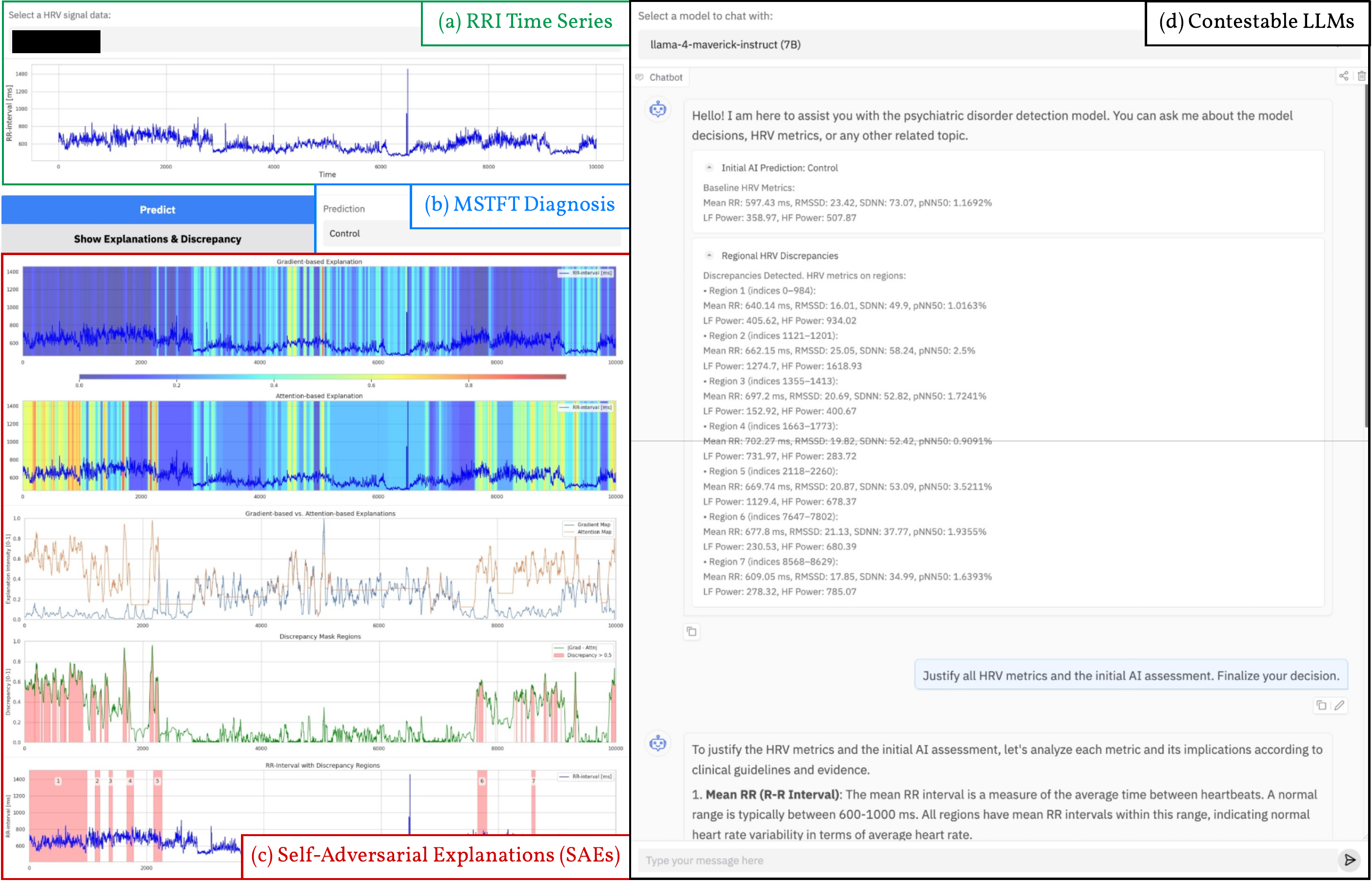}
    \caption{Contestable Diagnosis Interface (CDI) including: (a) raw RRI time series viewer, (b) psychiatric disorder diagnosis by MSTFT, (c) Self-Adversarial Explanations (SAEs) with gradient-based explanation, attention-based explanation, regions of discrepancy, and (d) contestable LLMs as a chatbot interface.}
    \label{fig:CDI}
\end{figure}

\subsubsection{Self-Adversarial Explanations (SAEs)}
\begin{figure}
    \centering
    \includegraphics[width=0.8\linewidth]{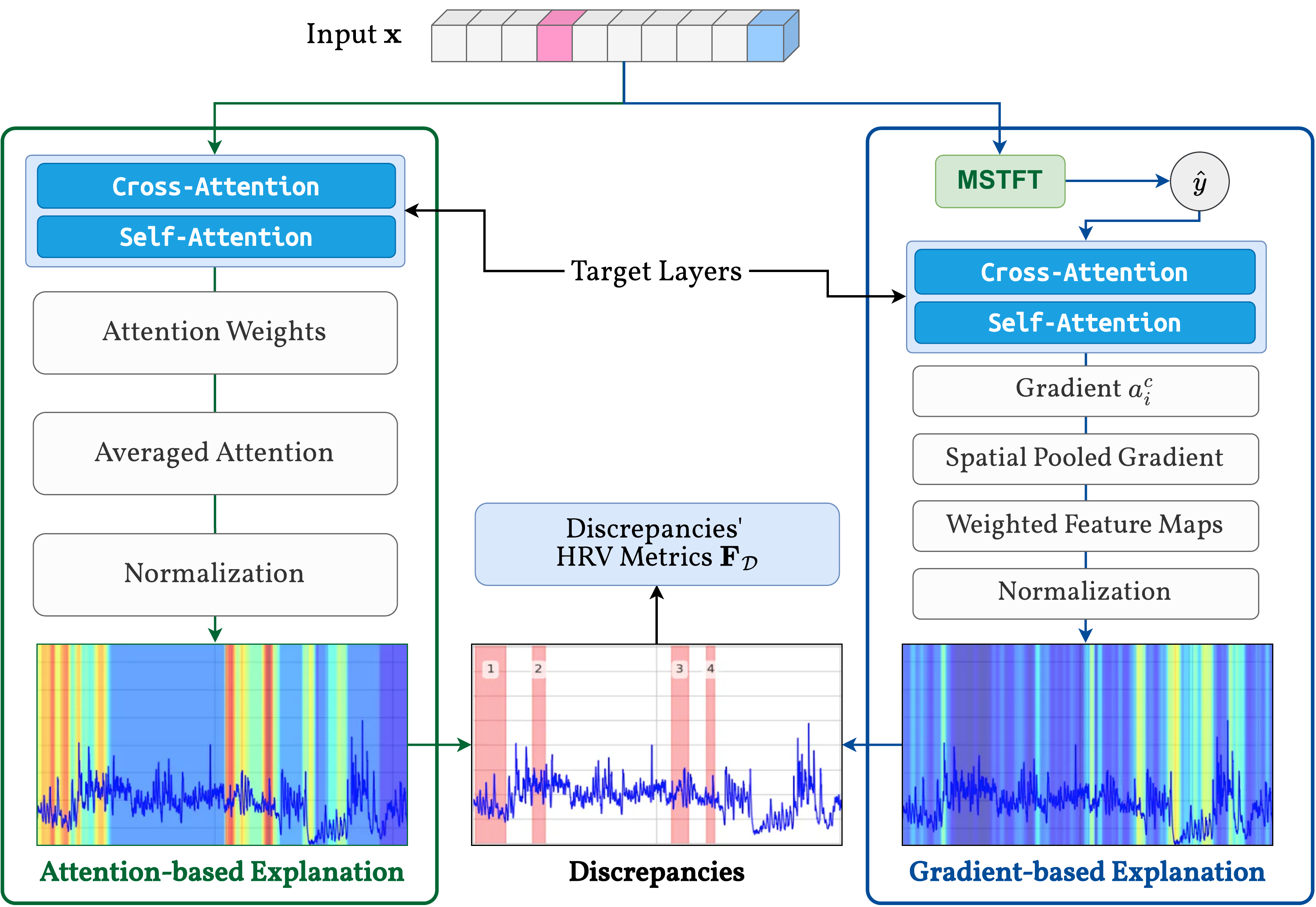}
    \caption{Self-Adversarial Explanations (SAEs).}
    \label{fig:advxai}
\end{figure}

As the core built-in safeguards of the contestable system, SAEs are designed to detect the discrepancies between attention-based and gradient-based explanation maps, then identify potential inconsistencies in the MSTFT model's decision-making process. 
As depicted in Figure~\ref{fig:advxai}, our SAEs approach identifies discrepancies by comparing two fundamentally different explanation methods: \textit{attention-based explanation} reveals where the model focuses during inference, and \textit{gradient-based explanation} highlights features with the strongest influence on the model's final prediction.
Discrepancies between these methods can offer valuable insights into model faithfulness, i.e., when they agree, the model likely makes decisions based on clinically relevant features; otherwise, when they diverge significantly, the model may be attending to regions that do not ultimately contribute to correct classification \cite{jain2019attention,liu2022rethinking}.

\paragraph{Attention-based Explanation} 
Attention-based explanations leverage the attention weights from transformer layers (i.e., cross-attention and self-attention) to identify regions in the input that contribute most to the model's predictions.
Given a sequence $\mathbf{x} = (\mathbf{x}_1, \mathbf{x}_2, \dots, \mathbf{x}_T)$, we extract attention weights from target layers defined as $\mathcal{L}$.
For each layer $l \in \mathcal{L}$, we compute the attention map $\mathbf{A}^{(l)}$ by averaging across all attention heads:
\begin{equation} 
    \mathbf{A}^{(l)} = \frac{1}{H} \sum_{h=1}^{H} \mathbf{A}_h^{(l)} 
\end{equation}
where $H$ is the number of attention heads and $\mathbf{A}_h^{(l)}$ is the attention weight matrix from the $h$-th head in layer $l$.
The combined attention-based explanation map $\mathbf{E}_{\text{attn}}$ is generated by averaging across all target layers:
\begin{equation} 
    \mathbf{E}_{\text{attn}} = \frac{1}{|\mathcal{L}|} \sum_{l \in \mathcal{L}} \mathbf{A}^{(l)} 
\end{equation}
To adapt the attention map to the original signal length, we employ an expansion function $\mathcal{E}$ that projects the sequence-level explanations to the time-domain:
\begin{equation} 
    \mathbf{E}_{\text{attn}}^T = \mathcal{E}(\mathbf{E}_{\text{attn}}, T) 
\end{equation}
where $T$ is the original signal length, and $\mathcal{E}$ performs a weighted distribution of attention values across overlapping segments. The final expanded map is normalized using z-score standardization followed by min-max scaling to ensure values lie within $[0,1]$.

\paragraph{Visual Gradient-based Explanation}
The gradient-based visual explanation employs the class activation mapping (CAM)-based approach to identify regions of high importance for the model's prediction.
For each layer $l \in \mathcal{L}$, we compute the gradient of the model's output with respect to the layer's activation:
\begin{equation} 
    \mathbf{G}^{(l)} = \frac{\partial \mathbf{y}_c}{\partial \mathbf{F}^{(l)}},
\end{equation}
where $\mathbf{y}_c$ is the model's prediction for the target class $c$ and $\mathbf{F}^{(l)}$ is the activation output of layer $l$.
The gradient weights are globally averaged to obtain importance weights $\alpha_k^{(l)}$ for each feature map $k$ in layer $l$:
\begin{equation} 
    \alpha_k^{(l)} = \frac{1}{Z} \sum_{i=1}^{Z} \frac{\partial \mathbf{y}_c}{\partial \mathbf{F}_{i,k}^{(l)}},
\end{equation}
where $Z$ is the spatial dimension of the feature maps. The gradient-weighted activation map for layer $l$ is then computed as:
\begin{equation} 
    \mathbf{L}_{\text{grad}}^{(l)} = \text{ReLU}\left(\sum{k} \cdot \alpha_k^{(l)} \cdot \mathbf{F}^{(l)}_k\right).
\end{equation}
The combined gradient-based explanation map $\mathbf{E}_{\text{grad}}$ is generated by averaging across all target layers:
\begin{equation} 
    \mathbf{E}_{\text{grad}} = \frac{1}{|\mathcal{L}|} \sum_{l \in \mathcal{L}} \mathbf{L}_{\text{grad}}^{(l)} 
\end{equation}
Similar to the attention-based explanation, we expand the gradient-based explanation to match the original signal length and normalize it to ensure values lie within $[0,1]$:
\begin{equation} 
\mathbf{E}_{\text{grad}}^{T} = \mathcal{E}(\mathbf{E}_{\text{grad}}, T) 
\end{equation}

\paragraph{Discrepancies Detection}
The self-adversarial nature of our approach emerges from comparing the two different explanation methods. 
To ensure meaningful comparison, we first align the attention-based explanation with the gradient-based explanation using Dynamic Time Warping (DTW):
\begin{equation} 
\mathbf{E}_{\text{attn}}^{T} = \text{DTW}(\mathbf{E}_{\text{attn}}^T, \mathbf{E}_{\text{grad}}^T) 
\end{equation}
The discrepancy map $\mathbf{D}$ is calculated as the absolute difference between the aligned attention-based explanation and the gradient-based explanation:
\begin{equation} 
    \mathbf{D} = |\mathbf{E}_{\text{attn}}^T - \mathbf{E}_{\text{grad}}^T| 
\end{equation}
Regions of high discrepancy ($\mathbf{D} > \rho$, where $\rho$ is a threshold parameter, empirically $\rho=0.5$) indicate areas where the two explanation methods disagree, highlighting potentially ambiguous or complex patterns in the data that require further human investigation. We define the discrepancy mask $\mathbf{M}_{\text{disc}}$ as:
\begin{equation} 
    \mathbf{M}_{\text{disc}}(t) = 
    \begin{cases} 1, & \text{if } \mathbf{D}(t) > \rho \\ 0, & \text{otherwise} \end{cases} 
\end{equation}
To identify contiguous regions of discrepancy, we merge adjacent positive mask values that are within a gap tolerance $\delta$:
\begin{equation} 
    \mathcal{D} = {(s_1, e_1), (s_2, e_2), \ldots, (s_n, e_n)} 
\end{equation}
where each pair $(s_i, e_i)$ represents the start and end indices of a contiguous region of discrepancy.

\paragraph{HRV Metrics Calculation}
For each discrepancy region $r_i = (s_i, e_i) \in \mathcal{D}$, we extract time-domain and frequency-domain HRV features, namely $\mathbf{F}_\mathcal{D}$. Time-domain features include:
\begin{gather} 
    \text{Mean RR}{[r_i]} = \frac{1}{e_i - s_i + 1} \sum{t=s_i}^{e_i} \mathbf{X}(t) \\ \text{RMSSD}{[r_i]} = \sqrt{\frac{1}{e_i - s_i} \sum{t=s_i}^{e_i-1} (\mathbf{X}(t+1) - \mathbf{X}(t))^2} 
    \\ 
    \text{SDNN}{[r_i]} = \sqrt{\frac{1}{e_i - s_i + 1} \sum{t=s_i}^{e_i} (\mathbf{X}(t) - \text{Mean RR}{[r_i]})^2} 
    \\ 
    \text{pNN50}{[r_i]} = \frac{100}{e_i - s_i} \sum_{t=s_i}^{e_i-1} \mathbf{1}(|\mathbf{X}(t+1) - \mathbf{X}(t)| > 50) 
\end{gather}
where $\mathbf{1}(\cdot)$ is the indicator function. Frequency-domain features include LF power (0.04-0.15 Hz) and HF power (0.15-0.40 Hz) estimated using Welch's method:
\begin{gather} 
    \text{Power Spectral Density: } P_{xx}(f) = \frac{1}{K} \sum_{k=0}^{K-1} \left| \sum_{t=0}^{L-1} w(t) \mathbf{X}_{r_i}(t + kR) e^{-j2\pi ft} \right|^2 
    \\
    \text{LF Power}{[r_i]} = \int_{0.04}^{0.15} P_{xx}(f)df 
    \\
    \text{HF Power}{[r_i]} = \int_{0.15}^{0.40} P_{xx}(f)df 
\end{gather}
where $P_{xx}(f)$ is the estimate of the power spectral density at frequency $f$, $w(t)$ is a window function, $L$ is the window length, $R$ is the window shift, and $K$ is the number of windows.

\subsubsection{Contestable LLMs}
We employ recent members of the open-sourced LLMs family (e.g., $\texttt{llama-4-maverick} \allowbreak \texttt{-instruct (7B)}$ \cite{meta2025llama}, $\texttt{phi-4-mini-instruct (3.8B)}$ \cite{abouelenin2025phi}, and $\texttt{gemma-3-instruct (27B)}$ \cite{team2025gemma}) as the core contestable LLMs due to their competitive performance on several datasets from the medical domain \cite{liang2023holistic}.
Each LLM processes a designed prompt template (see Template~\ref{template:A}, and receives identical inputs for each case, consisting of: patient profile information (when available), initial baseline MSTFT prediction $\hat{y}$ (``control'' or ``treatment''), complete HRV metrics extracted from the raw RRI time series $\mathbf{F}_\mathcal{R}$, and HRV metrics of all detected discrepancy regions $\mathbf{F}_\mathcal{D}$ (when present) to access the diagnosis information and deliver conversational explanations.
The generated explanations provide a concise and intuitive summary of psychiatric disorder detection, allowing clinicians to understand the model's rationale behind its decisions and to contest the faithfulness of the model's decisions.

\newtcolorbox[auto counter]{prompttemplate}[1][]{
  enhanced,
  fonttitle=\scshape,
  #1
}
\definecolor{ForestGreen}{RGB}{34,139,34}
\begin{figure*}[t!]
\begin{prompttemplate}[label=template:A,
  title={Template \thetcbcounter: Prompt template for Contestable LLMs}
]
\textbf{SYSTEM MESSAGE:} 
You are a helpful clinical decision support AI for psychiatric disorder (schizophrenia/bipolar disorder) diagnosis. Always:
\begin{enumerate}
    \item Think step-by-step before responding.
    \item Justify prior AI prediction and the interpretation of HRV metrics, referencing clinical guidelines or evidence when possible.
    \item  When the finalization request is queried, you must finalize the decision (only answer ``control'' or ``treatment''), but you may overturn prior AI prediction if, after reviewing all evidence, you are confident a different answer is correct. Clearly state the reason for any change.
    \item Provide accurate, current information using clinical guidelines.
    \item Avoid assumptions. Only use the provided data.
    \item Cross-validate findings with multiple sources.
    \item Flag urgent concerns immediately.
    \item Reference sources for non-standard conclusions.
    \item Maintain clarity with concise responses.
\end{enumerate}
\textbf{USER MESSAGE:}
\begin{enumerate}
    \item Patient profile (optional): \{\textcolor{ForestGreen}{age}\}, \{\textcolor{ForestGreen}{sex}\}.
    \item Prior AI Prediction: \{\textcolor{ForestGreen}{pd\_prediction}\}.
    \item Baseline HRV metrics: \{\textcolor{ForestGreen}{rri\_hrv\_metrics}\}.
    \item Regional HRV discrepancies: \{\textcolor{ForestGreen}{[region\_1\_metrics],[region\_2\_metrics],...,[region\_n\_metrics]}\}.
\end{enumerate}
\end{prompttemplate}
\end{figure*}

%% file: sec/5_model_exp.tex
\section{Experiment}\label{sec:exp}
To evaluate the effectiveness of our Heart2Mind framework, we conducted comprehensive experiments focusing on two key aspects: the diagnostic performance of the MSTFT model against state-of-the-art baselines, and the capability of our contestable LLMs system to validate correct predictions and identify potential errors. Our experimental evaluation employed the HRV-ACC dataset \cite{ksikazek10hrv} and implemented standard validation protocols to ensure reliable assessment of our proposed approach.

\subsection{Dataset}
We employed the HRV component in the HRV-ACC dataset \cite{ksikazek10hrv}, which stores raw RRI time series of 60 participants. The dataset includes the following participants: 30 diagnosed with schizophrenia/BP (labeled as ``treatment''/positive) and 30 controls (labeled as ``control''/negative). This dataset is considered balanced for training the model.
Each participant contributed 1.5–2 hours (minimum 70 minutes) of ECG recordings using a wearable Polar H10 chest-strap sensor, during which they followed light free-living protocols that included short corridor walks interleaved with seated rest periods.
For model training, we transformed each RRI series into a set of overlapping input sequences
\begin{equation}
    \mathbf{x}_i = (\mathbf{x}_{i_1}, \mathbf{x}_{i_2}, \dots, \mathbf{x}_{i_T}),\quad T=300
\end{equation}
producing \(N-T+1\) sequences from an original length \(N\). Consecutive sequences share \(T-1\) RRIs, i.e. \(\mathbf{x}_i\cap\mathbf{x}_{i+1}=\{\mathbf{x}_{i_2},\dots,\mathbf{x}_{i_T}\}\); thus every inner interval \(\mathbf{x}_k\) appears in exactly \(T\) contexts \(\{\mathbf{x}_{k-T+1},\dots,\mathbf{x}_k\}\).
This sliding-window strategy exposes the network to diverse local temporal neighborhoods, enhancing its ability to capture short- and mid-range autonomic dynamics. Before windowing, we rescaled each participant’s RRI signal to zero mean and unit variance to remove inter-subject scale differences. The resulting sample distribution remains balanced (30 positive vs 30 negative) and fully anonymized.

\begin{figure}[b!]
    \centering
    \begin{subfigure}[b]{0.43\linewidth}
        \centering
        \includegraphics[width=\textwidth]{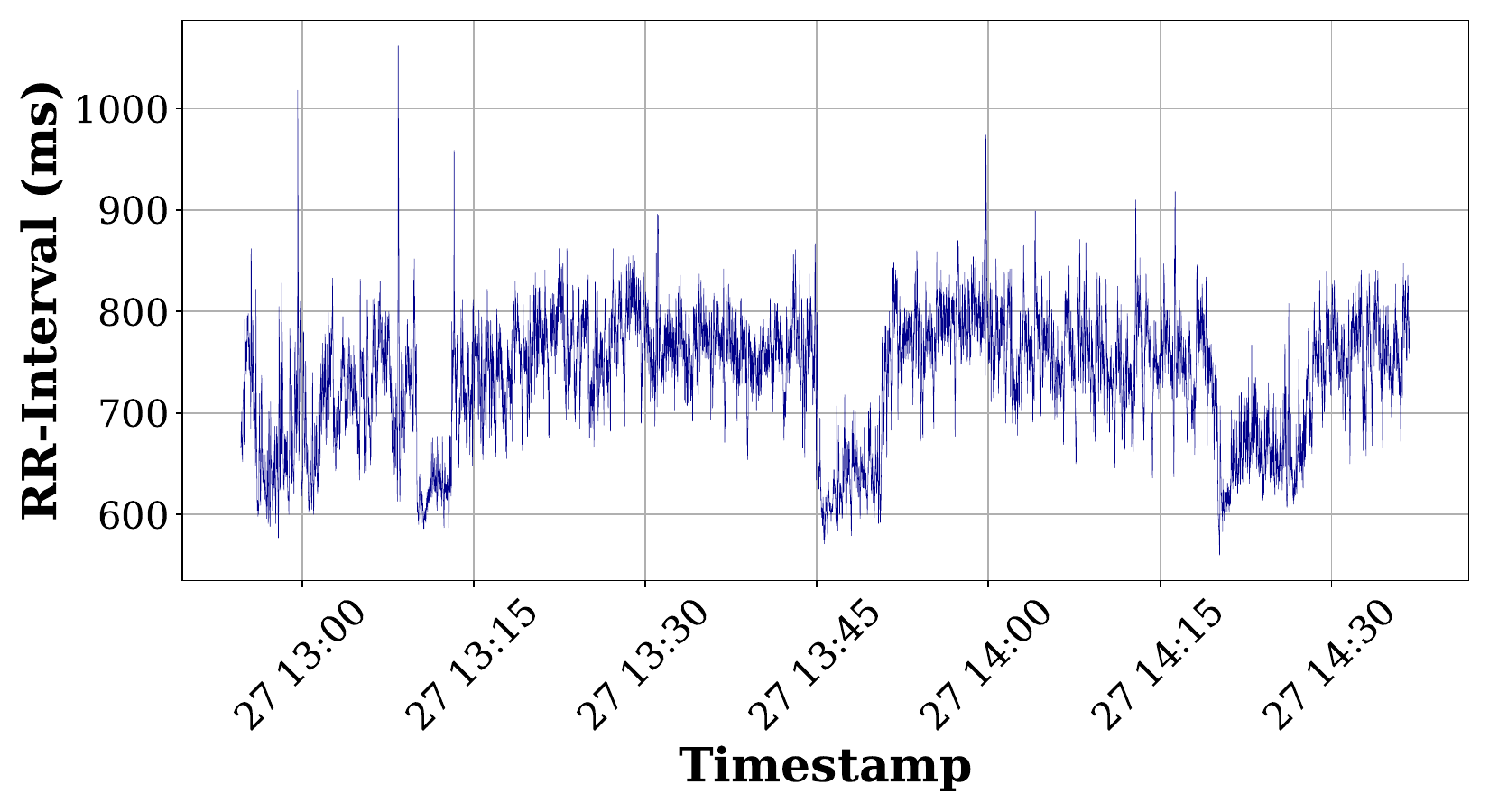}
        \caption{Control}
        \label{fig:control}
    \end{subfigure}
    \begin{subfigure}[b]{0.43\linewidth}
        \centering
        \includegraphics[width=\linewidth]{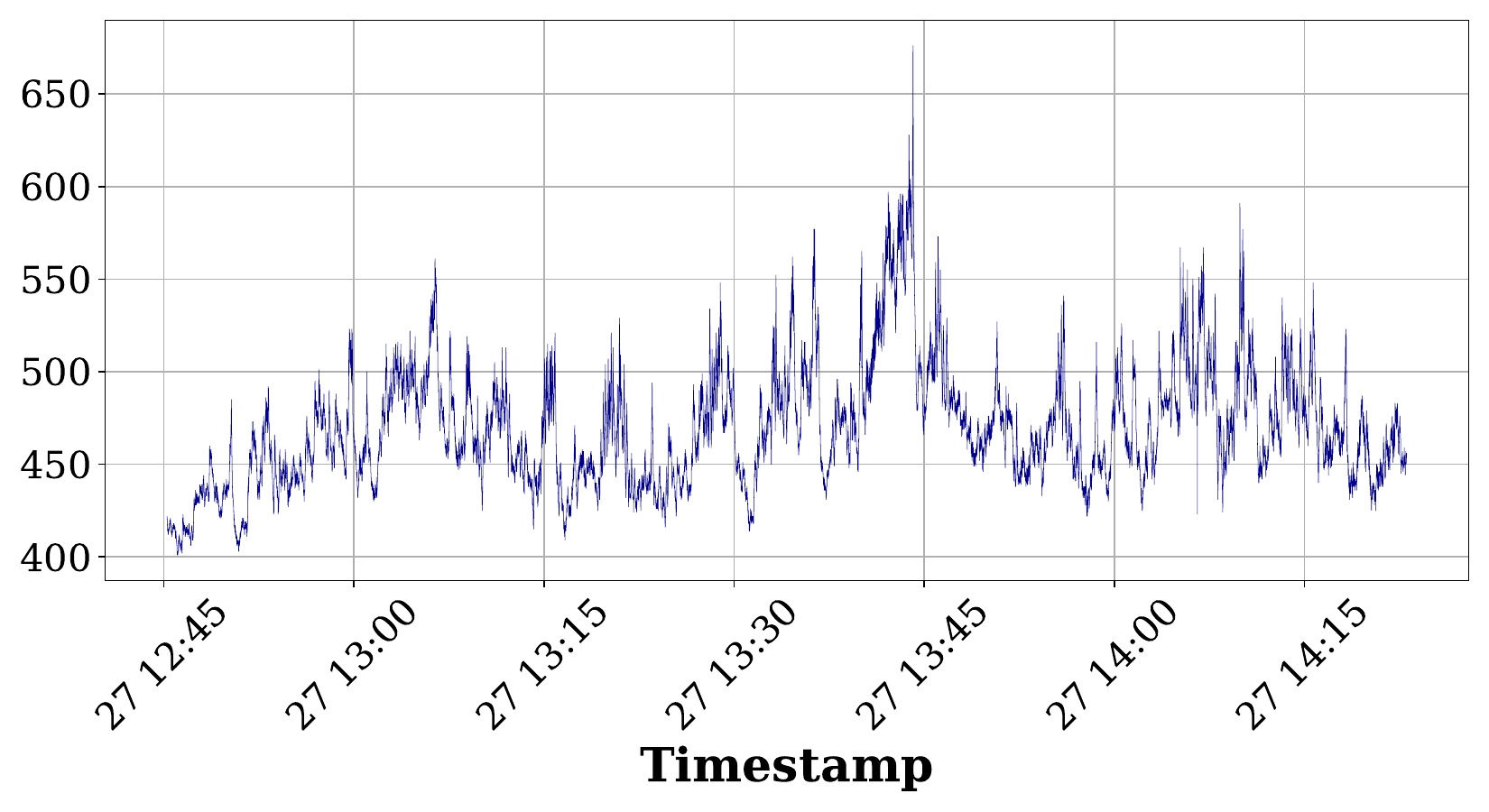}
        \caption{Treatment}
        \label{fig:treatment}
    \end{subfigure}
    \caption{Data samples from the (a) ``control'' and (b) ``treatment'' (schizophrenia/bipolar disorder) groups in the HRV-ACC dataset.}
    \label{fig:samples}
\end{figure}

\subsection{MSTFT Performance Evaluation}
This assessment examines the performance of our proposed MSTFT model in comparison with several baseline approaches. The evaluation was conducted using multiple performance metrics across two validation protocols, providing comprehensive insights into the model's effectiveness for diagnostic applications.

\subsubsection{Experimental Setup}

\begin{table}[]
    \centering
    \caption{Optimized set of hyperparameters for MSTFT.}
    \begin{tabular}{lclc}
    \toprule
       \textbf{Hyperparameter}  & \textbf{Value} & \textbf{Hyperparameter}  & \textbf{Value} \\
    \midrule
       Gaussian noise ($\lambda^2$)  & 0.1 & Cross-attention projection dimension ($d_p$) & 1024 \\
       Positional encoding wavelength ($\tau$) & $10^4$ & Attention key dimension ($d_k$) & 512 \\
       Convolutional embedded dimension ($d$) & 64 & Number of attention heads ($h$) & 16 \\
       Number of temporal blocks ($n_t$) & 2 & Augmentation strength ($\alpha$) & 0.3 \\
       Stochastic skip survival probability ($p_s$) & 0.8 & Learning rate ($\eta) $ & 1e-5 \\ 
       Number of frequency blocks ($n_f$) & 2 & Weight decay & 1e-5 \\
    \bottomrule
    \end{tabular}
    \label{tab:mstft_best_hyper}
\end{table}

To comprehensively evaluate our proposed MSTFT model, we compared it against several strong baselines: (1) backbone variants of our architecture, i.e., 1D-CNN and Transformer, and (2) state-of-the-art approaches for psychiatric disorders detection, i.e., the attention-guided method of \citet{misgar2024unveiling}, the convolutional nearest neighbor approach of \citet{buza2023simple}, and the ensemble of SVMs with GRU-based NNs of \citet{ksikazek2025deep}.
Our implemented models and attention-guided method of \citet{misgar2024unveiling} were tuned by randomized hyperparameter search on the training portion of each split, with 20\% of that portion reserved as an inner validation set. 
Table~\ref{tab:mstft_best_hyper} presents the optimal hyperparameter configuration for our MSTFT model.
For the models of \citet{buza2023simple} and \citet{ksikazek2025deep}, we directly report the performance metrics from their original publications to ensure fair comparison.
All methods were evaluated using a comprehensive set of metrics, i.e., Accuracy, Precision, Recall, F1 score, and ROC-AUC, across two standard validation protocols:
\begin{itemize}
    \item \textbf{5-fold cross-validation:} 48 participants for training and 12 for testing in every fold. The final results are averaged across folds.  
    \item \textbf{Leave-one-out cross-validation (LOOCV):} We employed LOOCV because the clinical datasets (e.g., our employed HRV-ACC dataset) are typically small, heterogeneous, and highly sensitive to inter-individual variability. Each model is trained on 59 participants and tested on a single held-out subject, then repeated 60 times, and the final results are averaged.
\end{itemize}

\begin{table*}[tb]
    \centering
    \caption{The performance results are presented with the highest values indicated in \textbf{bold}, and the second-highest values \underline{underlined}. ``-'' metrics are not available in the original work.}
    \label{tab:perf}
    \begin{tabular}{lccccc}
        \toprule
            \textbf{Model} & \textbf{Accuracy} & \textbf{Precision} & \textbf{Recall} & \textbf{F1 Score} & \textbf{AUC}  \\
        \midrule
            \multicolumn{6}{c}{\textbf{(a) 5-fold cross-validation}} \\
            1D-CNN & 0.750  & 0.747& 0.826 & \underline{0.784} & 0.895 \\
            Transformer & 0.750 & 0.710  & \underline{0.833} & 0.767 & 0.875  \\
            \citet{misgar2024unveiling} & 0.766 & \underline{0.833} & 0.667 & 0.741 & \underline{0.928}  \\
            \citet{ksikazek2025deep} & \underline{0.830} & - & - & - & - \\
            MSTFT (Ours) & \textbf{0.891}  & \textbf{0.877} & \textbf{0.903} & \textbf{0.890} & \textbf{0.961} \\
        \midrule
            \multicolumn{6}{c}{\textbf{(b) Leave-one-out cross-validation}} \\
            1D-CNN & 0.800 & 0.781 & \underline{0.833} & 0.806 & 0.826 \\
            Transformer & \underline{0.850} & \underline{0.920} & 0.767 & \underline{0.836} & 0.850 \\
            \citet{misgar2024unveiling} & 0.833 & 0.884 & 0.767 & 0.836 & 0.906 \\
            \citet{buza2023simple} & 0.833 & - & - & - & \underline{0.910}  \\
            \citet{ksikazek2025deep} & 0.800 & - & - & - & - \\
            MSTFT (Ours) & \textbf{0.917} & \textbf{0.963} & \textbf{0.867} &  \textbf{0.913} & \textbf{0.940} \\
        \bottomrule
    \end{tabular}
\end{table*}

\subsubsection{Results on 5-fold cross-validation}
Table \ref{tab:perf}a shows that MSTFT achieved the best score on every metric reported.
Its accuracy of 0.891 was 6.1\% above the strongest baseline (0.830 by the model of \citet{ksikazek2025deep}), 12.5\% above the re-implemented model of \citet{misgar2024unveiling} (0.766), and 14.1\% beyond the 1D-CNN.
The significant gain in recall (0.903) indicated that MSTFT correctly identified almost all people with schizophrenia/bipolar disorder, an essential property in clinical screening without sacrificing precision (0.877).
Among the backbone components, 1D-CNN and Transformer reached the same accuracy (0.750) but showed different error profiles, where Transformer maximized recall (0.833), but came at the cost of a markedly lower precision of 0.710. In contrast, 1D-CNN was slightly more balanced, suggesting that temporal convolutions alone could not capture the full variability present in the segments. 

By contrast, MSTFT raised precision to 0.877, producing the largest harmonic mean, \(F_{1}=0.890\), which was 0.106 higher than the best baseline value (0.784 from 1D-CNN). MSTFT's AUC of 0.961 exceeded other models and indicates that MSTFT preserved a large safety margin over a broad range of decision thresholds. 
These gains suggested that MSTFT benefited from the joint time–frequency representation, allowing it to discriminate subtle rhythm alterations that elude purely temporal or convolutional encoders.

\subsubsection{Results on Leave-one-out cross-validation} 
Under the more rigid LOOCV protocol, as shown in Table \ref{tab:perf}b, our MSTFT maintained its superior performance across metrics.
Our model achieved an accuracy of 0.917, outperforming the next best model (Transformer at 0.850) by 6.7 percentage points. The precision score of 0.963 was particularly impressive, indicating that fewer than 4\% of health controls were wrongly flagged as positive.
The recall value of 0.867 ensured that most true cases were successfully detected, while the F1 score of 0.913 confirmed the model's balanced performance. The AUC of 0.940 further demonstrated robust classification performance across different threshold settings.
In this protocol, Transformer emerged as the second-best model with strong precision (0.920) but weaker recall (0.767) compared to MSTFT. 
In contrast, methods of \citet{misgar2024unveiling} and \citet{buza2023simple} showed competitive AUC values (0.906 and 0.910, respectively) but fell short in overall accuracy and F1.

In general, our proposed MSTFT significantly advanced state-of-the-art classification performance, consistently outperforming baseline approaches across all metrics in both validation protocols. Integrating complementary components created a powerful architecture where 1D-CNN extracted local temporal and frequency features, and Transformer captured global relationships. This joint time-frequency representation proved essential because physiological signals contain critical information distributed across both domains. Temporal feature extractor alone missed frequency patterns indicating subtle rhythm alterations, while frequency analysis without temporal context lost sequential relationships needed for accurate diagnosis. The multi-scale approach in MSTFT further enabled detection of patterns at different timescales, capturing both rapid transitions and slow-evolving trends invisible to single-scale methods. 

\subsection{Analysis of Contestable LLMs with SAEs}
In this section, we conducted a comprehensive evaluation of contestable LLMs with SAEs to create a contestable diagnosis system for psychiatric disorders. This analysis addressed two challenges in clinical AI systems: (1) the need for transparency in model decision-making; (2) the ability to identify and correct potential errors.
We first evaluated the effectiveness of SAEs in detecting regions of model inconsistency, then assessed how well different LLMs can leverage these identified discrepancies to validate correct predictions and contest incorrect ones.

\begin{table}[h!]
    \centering
    \caption{Number of classification cases by baseline MSTFT and contestable LLMs supported by SAEs. The arrows ($\uparrow$/$\downarrow$) indicate the higher/lower, the better.}
    \begin{tabularx}{\linewidth}{lXXXXXX}
    \toprule
        Model & \textcolor{NavyBlue}{Retain (TP) $\uparrow$} & \textcolor{NavyBlue}{Retain (TN) $\uparrow$} & \textcolor{BrickRed}{Overturn (TP/TN) $\downarrow$} & \textcolor{ForestGreen}{Overturn (FN) $\uparrow$} & \textcolor{ForestGreen}{Overturn (FP) $\uparrow$} & \textcolor{BrickRed}{Retain (FN/FP) $\downarrow$} \\
    \midrule
        $\texttt{llama-4-maverick-instruct(7B)}$ & 27 & 27 & 0 & 0 & 1 & 5 \\
        $\texttt{phi-4-mini-instruct(3.8B)}$ & 27 & 27 & 0 & 1 & 0 & 5 \\
        $\texttt{gemma-3-instruct(27B)}$ & 27 & 27 & 0 & 2 & 1 & 3 \\
    \bottomrule
    \end{tabularx}
    \label{tab:perf_by_cat}
\end{table}
\begin{figure}[h!]
    \centering
    \includegraphics[width=.9\linewidth]{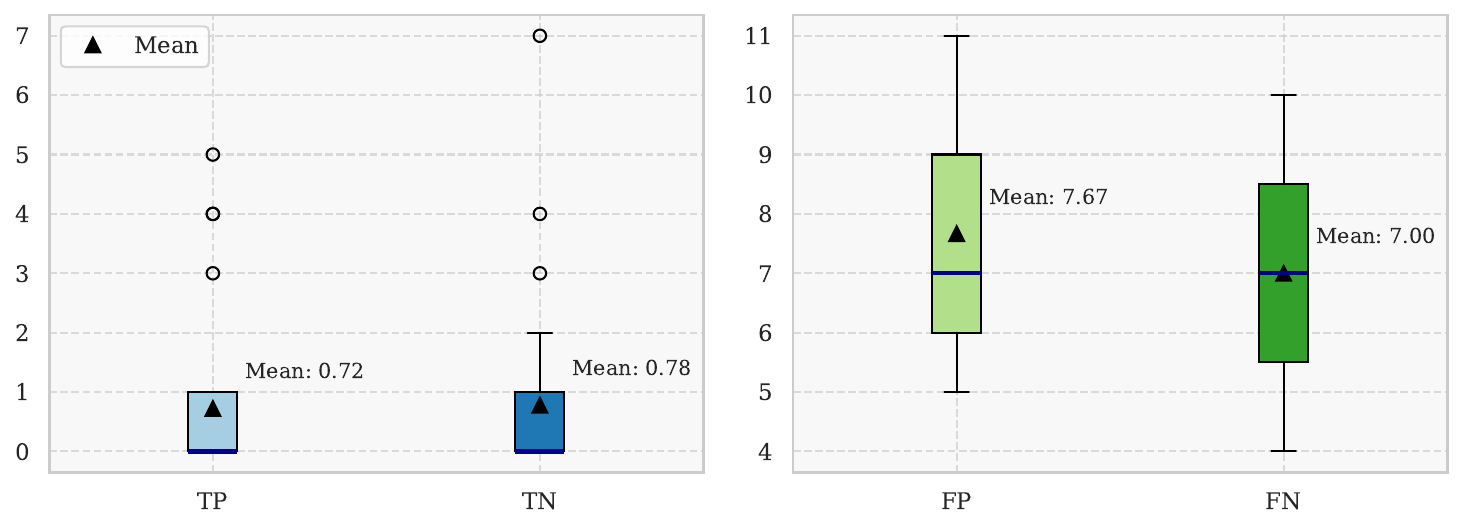}
    \caption{Number of discrepancies detected by SAEs for classification cases of the best MSTFT checkpoint across 5-fold cross-validation on the HRV-ACC dataset.}
    \label{fig:discrepancies_by_cat}
\end{figure}

\subsubsection{SAEs Evaluation}
We conducted an evaluation of SAEs to detect inconsistent regions in MSTFT predictions that may indicate unfaithful behavior. These discrepancies could warrant additional clinical scrutiny, as they represented areas where the model's decision-making process showed internal inconsistencies. 
For this evaluation, we used the best-performing MSTFT checkpoint across 5-fold cross-validation as our baseline model. 
This model achieved a balanced distribution of classification outcomes on the HRV-ACC dataset: 27 true positive (TP) cases, 27 true negative (TN) cases, 3 false negative (FN) cases, and 3 false positive (FP) cases.

Figure~\ref{fig:discrepancies_by_cat} presents a quantitative analysis of discrepancies between visual attention-based and gradient-based explanations detected by SAEs across different prediction categories.
We observed a strong contrast between correct and incorrect predictions.
In correctly classified cases (TP and TN), the two visual explanation methods demonstrated a strong alignment. TN cases
showed a mean discrepancy count of only 0.78, while TP cases exhibited an even lower average of 0.72. This alignment empirically confirmed that when MSTFT predicted correctly, its attention mechanisms focused on regions where the gradients were also significant for classification.
While some correctly classified cases contained a high number of discrepancies, with maximums of 5 in one TP case and 7 in one TN case, these represented outliers rather than the norm. Even in these cases with higher discrepancy counts, the model still reached the correct prediction, suggesting a degree of robustness in the decision-making process despite some internal inconsistencies. Furthermore, the TP case with 5 discrepancies was demonstrated as an example to challenge the ability of contestable LLMs in Section~\ref{sec:llm}.

Incorrect predictions (FP and FN) presented a markedly different pattern. FP cases demonstrated a substantially higher mean discrepancy count of 7.67 and the highest of 11, while FN cases averaged 7.0 and had a maximum of 10 discrepancies. This reflected an increase in discrepancies, indicating erroneous predictions, where MSTFT's attention focused on regions that did not meaningfully contribute to final decisions according to gradient explanations. The model exhibited unfaithful behavior by attending to features that led to misclassifications.

The consistent correlation between discrepancy count and prediction accuracy established a clear threshold effect: when discrepancies exceeded approximately 5 to 6 regions, prediction reliability became substantially compromised. This observation had important implications for clinical implementation. By monitoring the number of discrepancies between visual attention-based and gradient-based explanation methods, we can flag potentially unreliable predictions for additional review without requiring ground truth labels, effectively creating an uncertainty quantification mechanism that operates during inference time.

The ability to identify potentially erroneous predictions is valuable, but ideally, we would also have mechanisms to correct these errors or at least provide alternative interpretations for clinical consideration. This need led us to explore whether contestable LLMs can analyze the detected discrepancies and either validate or contest the initial MSTFT predictions based on their understanding of clinical patterns in HRV data.

\subsubsection{Contestable LLMs Performance}\label{sec:llm}
Having established the relationship between explanation discrepancies and model faithfulness, we investigated whether contestable LLMs can effectively use these discrepancies to contest or validate MSTFT predictions.

In this experiment, we implemented contestable LLMs using three state-of-the-art open-sourced LLMs with varying parameter sizes: $\texttt{llama-4-maverick-instruct (7B)}$, $\texttt{phi-4-mini-instruct (3.8B)}$, and $\texttt{gemma-3} \allowbreak \texttt{-instruct (27B)}$.
To ensure fair comparison between these architecturally different LLMs, we standardized generation parameters across all models: maximum token length is set to 2048, temperature (controlling variability and randomness of model's responses) to 0.8, and top P (controlling the randomness of the model's responses) to 0.1. 
Each LLM received identical inputs for each case, consisting of: patient profile information (when available), initial baseline MSTFT prediction (``control'' or ``treatment''), complete HRV metrics extracted from the raw RRI time series, and HRV metrics of all detected discrepancy regions (when present). 
We then prompted each model with a standardized query asking it to validate the initial MSTFT prediction and provide a final decision. Importantly, we intentionally avoided incorporating additional expert knowledge of specific medical guidelines in the prompt, challenging the model's inherent ability to interpret HRV metrics and reach valid clinical conclusions.

Our evaluation protocol of contestable LLMs examined two distinct cases as follows: (1) A TP case where MSTFT correctly classified the samples and the LLMs should ideally validate and \textcolor{NavyBlue}{retain} these decisions, and (2) A FN case where MSTFT made errors and the LLMs should ideally \textcolor{BrickRed}{overturn} and correct these misclassifications. 
These scenarios required different capabilities from the contestable LLMs: the first tested their ability to recognize valid patterns consistent with the initial prediction, while the second tested their capacity to identify inconsistencies and propose more accurate alternatives.

\paragraph{True Prediction Cases}
\begin{figure}[hb!]
    \centering
    \includegraphics[width=\linewidth]{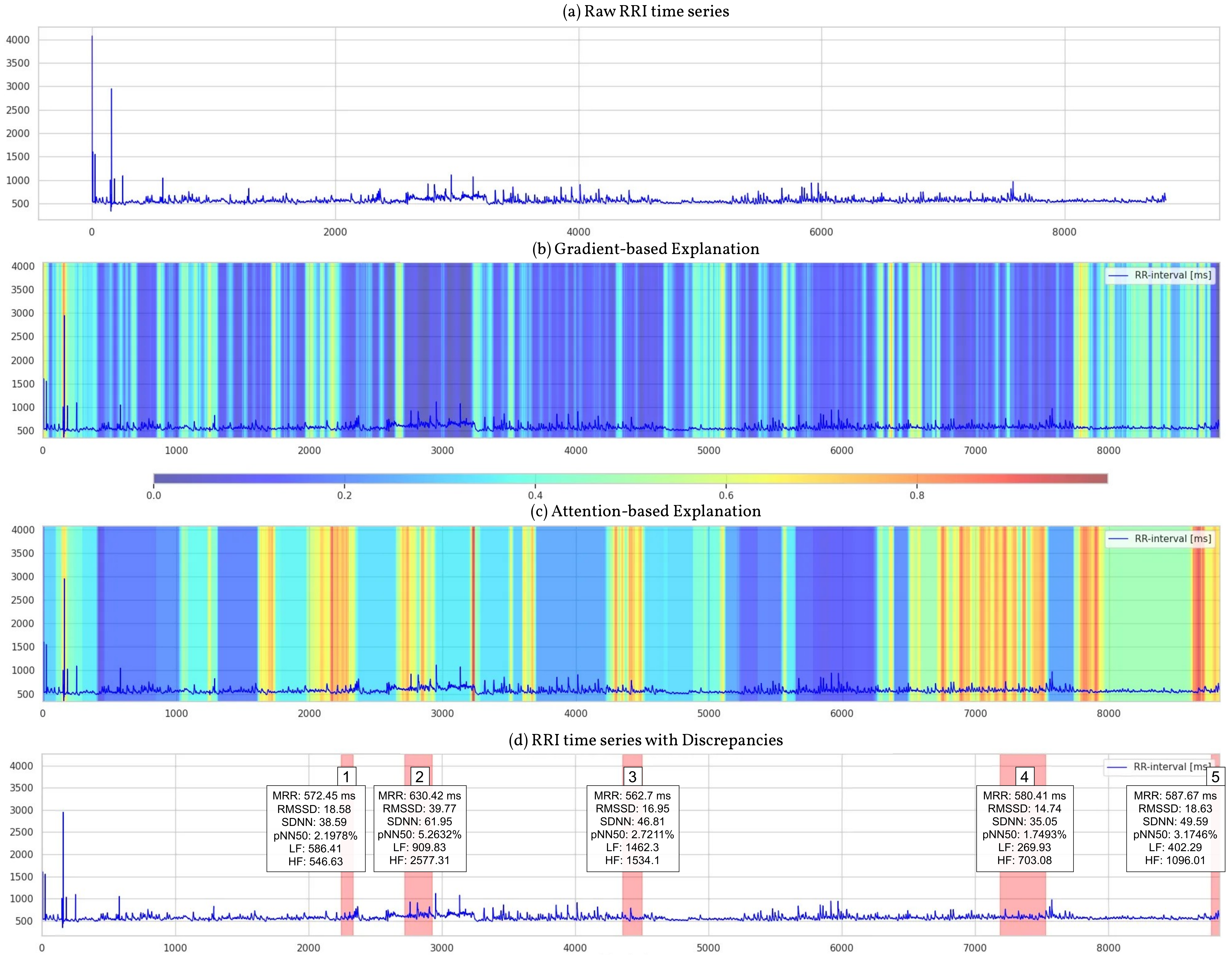}
    \caption{SAEs on a TP case (\textit{treatment\_40}): (a) Raw RRI time series, (b) Gradient-based Explanation, (c) Attention-based Explanation, and (d) 5 discrepancies detected by SAEs.}
    \label{fig:xai_tp}
\end{figure}
\begin{figure}[ht!]
    \centering
    \includegraphics[width=.95\linewidth]{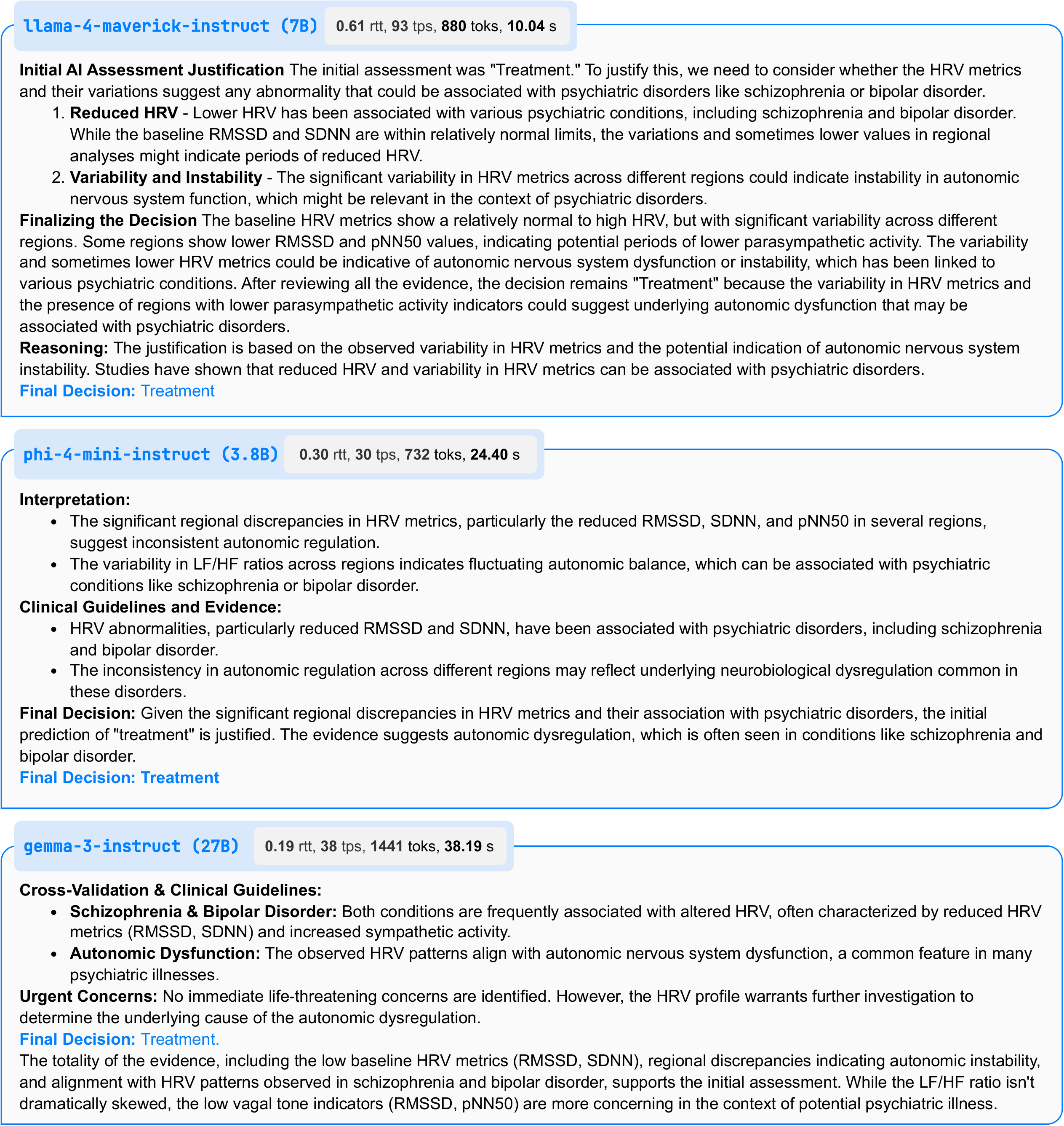}
    \caption{Response excerpts of contestable LLMs in a TP case. The final decision (in \textcolor{NavyBlue}{blue}) indicates the model \textcolor{NavyBlue}{retains} with the correct initial prediction of baseline MSTFT. Output metrics: time to first token (rtt), token per second (tps), number of output tokens (toks), output time (s).}
    \label{fig:tp}
\end{figure}

In cases where the baseline MSTFT made correct predictions (TP/TN), an ideal contestable system should validate and \textcolor{NavyBlue}{retain} the initial MSTFT decisions while providing clear clinical justification. 
Table~\ref{tab:perf_by_cat} presents the comprehensive results of our evaluation, showing that all three contestable LLMs successfully affirm and justify all correct predictions from the baseline MSTFT (27/27 TP and 27/27 TN cases).
This retention rate demonstrated the LLMs' ability to accurately interpret HRV patterns and metrics consistent with both treatment-requiring conditions and healthy controls.

This performance of contestable LLMs aligned with our theoretical expectations. Since correct MSTFT predictions exhibited few discrepancies, the corresponding HRV metrics presented a coherent clinical picture that LLMs can readily interpret. Our analysis showed that all three models consistently identify key diagnostic indicators in the HRV data, including RMSSD, SDNN, pNN50, and LF/HR ratio patterns that differentiate between psychiatric conditions and healthy states.

We conducted detailed case studies to better understand LLM reasoning processes. Figure~\ref{fig:tp} presents the responses of all three contestable LLMs for a particularly challenging TP case with 5 discrepancies (see Figure~\ref{fig:xai_tp}), which is the maximum observed in any correct prediction. Despite the high number of discrepancies, all three models independently reached agreement to \textcolor{NavyBlue}{retain} the initial ``treatment'' classification. Each provided specific rationales, citing relevant HRV metrics that influence their assessment:
\begin{itemize}
    \item $\texttt{llama-4-maverick-instruct (7B)}$ focused on ``significant variability across different regions'' and ``lower parasympathetic activity indicators''.
    \item $\texttt{phi-4-mini-instruct (3.8B)}$ identified on ``significant regional discrepancies in HRV metrics'' and ``variability in LF/HF ratios across regions''.
    \item $\texttt{gemma-3-instruct (27B)}$ focused on ``regional discrepancies indicating autonomic instability'' and ``low vagal tone indicators (RMSSD, pNN50)''.
\end{itemize}

Interestingly, the $\texttt{llama-4-maverick-instruct (7B)}$ model can output in the shortest time (10.04 seconds), significantly faster than the smallest $\texttt{phi-4-mini-instruct (3.8B)}$ model (24.40 seconds) despite having nearly twice the parameter count, which had important implications for real-time clinical applications where response latency matters.

\paragraph{False Prediction Cases}
\begin{figure}[b!]
    \centering
    \includegraphics[width=\linewidth]{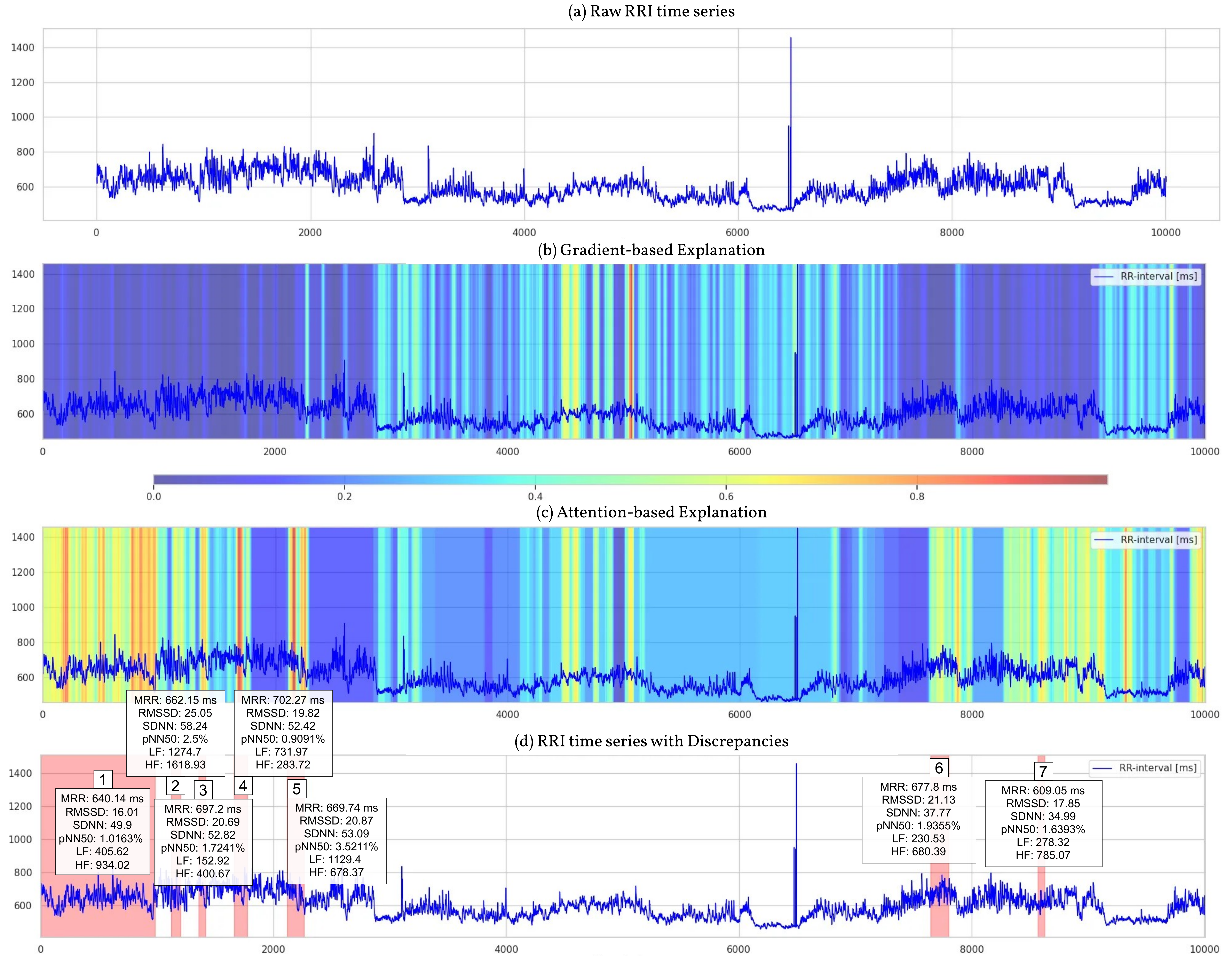}
    \caption{SAEs on a FN case (\textit{treatment\_1}): (a) Raw RRI time series, (b) Gradient-based Explanation, (c) Attention-based Explanation, and (d) 7 discrepancies detected by SAEs.}
    \label{fig:xai_fn}
\end{figure}
\begin{figure}[ht!]
    \centering
    \includegraphics[width=.95\linewidth]{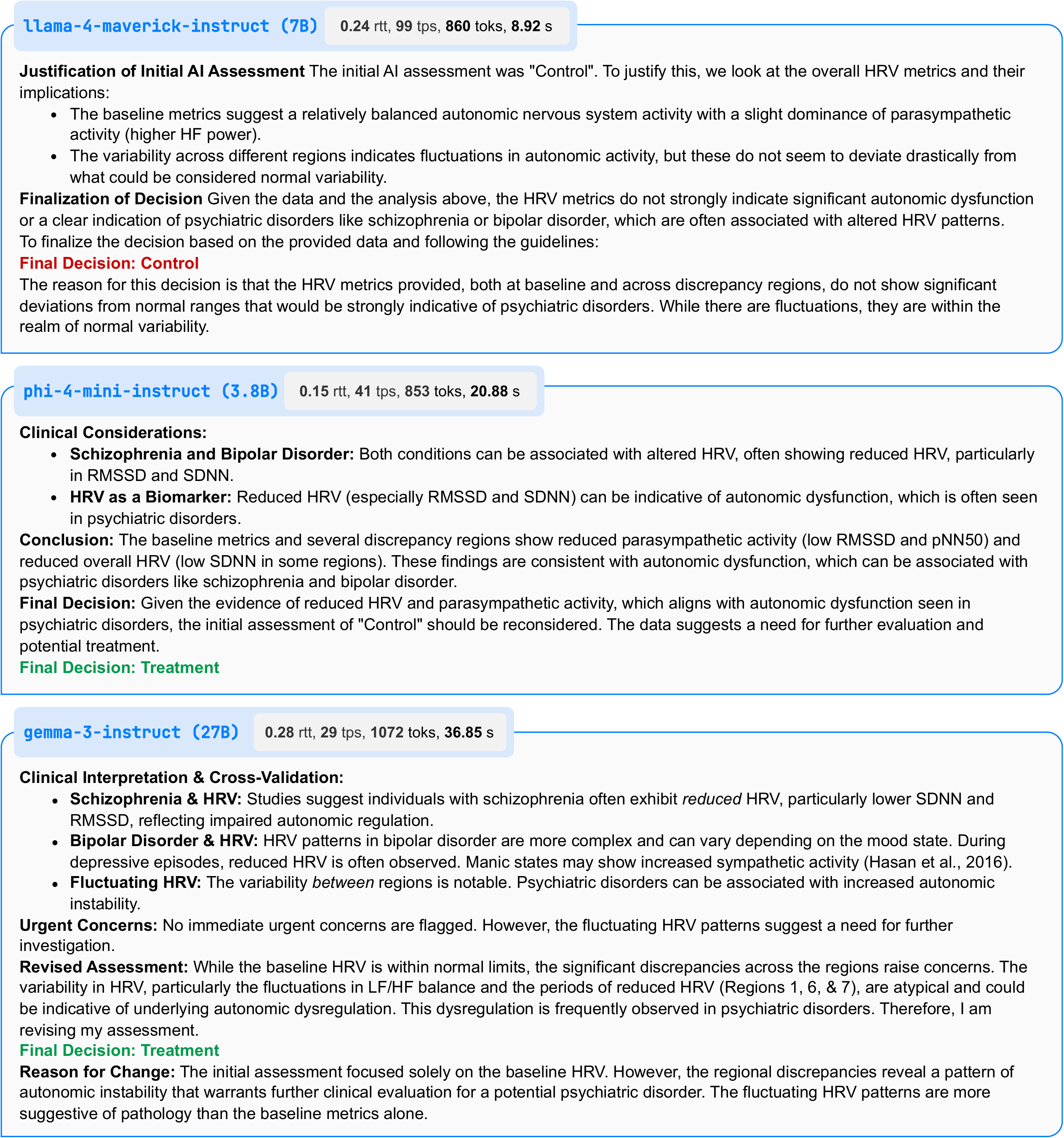}
    \caption{Response excerpts of contestable LLMs in a FN case. The final decision (in \textcolor{BrickRed}{red}) indicates the model \textcolor{BrickRed}{retains} with the incorrect initial prediction of baseline MSTFT. The final decision (in \textcolor{ForestGreen}{green}) indicates the model \textcolor{ForestGreen}{overturns} the incorrect initial prediction of baseline MSTFT to the correct prediction. Output metrics: time to first token (rtt), token per second (tps), number of output tokens (toks), output time (s).}
    \label{fig:fn}
\end{figure}

While validating correct predictions is important, the true test of contestable AI is its ability to identify and correct model errors. 
In cases where MSTFT made incorrect predictions (FN/FP), we evaluated whether contestable LLMs can $\textcolor{ForestGreen}{\text{overturn}}$ these errors and provide accurate classifications. This capability is particularly critical in clinical contexts, where FN could result in delayed treatment and FP might lead to unnecessary interventions.

Regarding the FN and FP cases in Figure~\ref{fig:discrepancies_by_cat}, despite the limited sample of incorrect predictions (3 FN and 3 FP cases), all three LLMs successfully contested at least one erroneous prediction.
The contestation effectiveness correlated strongly with model size, with the largest model $\texttt{gemma-3-instruct (27B)}$ showing the strongest performance by $\textcolor{ForestGreen}{\text{overturning}}$ 3 out of 6 incorrect predictions (2/3 FN cases and 1/3 FP cases).
The mid-sized $\texttt{llama-4-maverick-instruct (7B)}$ successfully $\textcolor{ForestGreen}{\text{overturned}}$ 1 FP case but failed to correct any FN cases. Interestingly, despite being the smallest model, $\texttt{phi-4-mini-instruct (3.8B)}$ successfully $\textcolor{ForestGreen}{\text{overturned}}$ 1 FN case. 
This pattern suggested that while larger models generally performed better at contestation, different architectures may have distinct strengths in analyzing various types of misclassifications.

To better understand the reasoning processes behind these contestation decisions, we conducted a detailed analysis of a specific case. Figure~\ref{fig:fn} displays the complete responses of all three contestable LLMs for a representative FN case with 7 discrepancies (see Figure~\ref{fig:xai_fn}), where the HRV metrics exhibited high perplexity and variability across different regions. Both $\texttt{gemma-3-instruct (27B)}$ and $\texttt{phi-4-mini-instruct (3.8B)}$ correctly identified autonomic dysregulation patterns indicative of psychiatric conditions and $\textcolor{ForestGreen}{\text{overturned}}$ the ``control'' prediction to ``treatment.'' Their analyses specifically highlighted:
\begin{itemize}
    \item $\texttt{phi-4-mini-instruct (3.8B)}$ identified on ``reduced parasympathetic activity (low RMSSD and pNN50)'' and ``reduced overall HRV (low SDNN in some regions)''.
    \item $\texttt{gemma-3-instruct (27B)}$ focused on ``pattern of autonomic instability'' and ``fluctuations in LF/HF balance and the periods of reduced HRV''.
\end{itemize}
In contrast, $\texttt{llama-4-maverick-instruct (7B)}$  $\textcolor{BrickRed}{\text{retained}}$ the incorrect ``control'' prediction. Its justification focused on different aspects: ``fluctuations are within the realm of normal variability.''
This divergence in interpretation highlighted significant differences in medical knowledge representation across LLMs. The larger models appeared to have more sophisticated representations of the relationship between autonomic dysregulation and psychiatric conditions, enabling more accurate contestation of FN cases.

In general, the observed performance patterns across both correct and incorrect predictions demonstrated the feasibility of using LLMs as contestable agents in clinical AI systems, especially in the context of psychiatric disorder diagnosis. Even without specialized medical fine-tuning, these models showed a remarkable ability to interpret complex HRV patterns and made clinically meaningful judgments. This suggested substantial potential for further improvements through domain-specific optimization and explicit incorporation of medical knowledge. 
Our analysis also revealed model-specific strengths and weaknesses. While $\texttt{gemma-3-instruct (27B)}$ showed the strongest overall contestation performance, it took significantly longer to generate responses (36.85-38.19 seconds) than the other models. $\texttt{llama-4-maverick-instruct (7B)}$ excelled at identifying FP but struggled with FN, while $\texttt{phi-4-mini-instruct (3.8B)}$ shows the opposite pattern.
These findings suggested that ensemble approaches using multiple LLMs might provide more robust contestation than reliance on any single model. Additionally, the superior performance of larger models indicated a clear pathway for ongoing improvement as more powerful LLMs become available, with potential for future models to achieve even higher contestation rates.

%% file: sec/7_dcs.tex
\section{Discussion}\label{sec:disc}
This section examines the broader implications of our Heart2Mind system, which successfully integrates wearable ECG technology with contestable AI for psychiatric disorder detection. Our contributions demonstrate that cardiac biomarkers can provide objective indicators for psychiatric conditions while maintaining essential human oversight in the diagnostic process. The discussion focuses on two key areas. First, we explore how wearable ECG devices show promise for continuous psychiatric monitoring through autonomic nervous system assessment, highlighting both their potential and limitations in clinical application. Second, we analyze our limitations, challenges, and opportunities in developing truly human-centered contestable AI systems for healthcare, including the significance of our SAEs and contestable LLMs system. Throughout this section, we examine how these technological advances could transform psychiatric care while addressing the ethical, clinical, and technical considerations necessary for responsible implementation.

\subsection{Potential of Wearable ECG Devices in Psychiatric Disorder Detection}
Wearable ECG devices are emerging as powerful tools in psychiatric care, especially for disorders like schizophrenia and bipolar disorder. 
These devices measure heart activity and HRV, offering insight into ANS function, which has been shown to be dysregulated in people with certain psychiatric disorders, such as bipolar disorder and schizophrenia. 
We found that both our explanation methods consistently highlight areas with rapid changes in the RRI time series, aligning with prior research showing lower HRV in people with schizophrenia and bipolar disorder compared to healthy controls \cite{faurholt2017heart,benjamin2021heart,ksikazek2025deep,ramesh2023heart,ksikazek2023analysis}. 
Also, schizophrenia and bipolar disorder are both associated with ANS dysregulation, show characteristic patterns of autonomic imbalance, and reflect reduced parasympathetic tone and elevated sympathetic activity.
Such autonomic dysfunction has been linked to symptom severity and illness progression in psychiatric conditions, making HRV a promising objective biomarker of mental state.
By continuously monitoring ECG signals, wearable devices can detect these subtle cardiac changes in real time, opening new avenues for early detection and monitoring of psychiatric disorders.

Although wearable devices tend to be the most cost-effective compared with other diagnostic and treatment monitoring methods in healthcare \cite{duarte2019lead,cullen2021experience}, this depends on the context of the disease we choose to monitor and the general availability of systemic healthcare \cite{ksikazek10hrv,velasquez2024economic}.
As wearables mostly rely on general physiological signals, they are not self-sufficient in their diagnostic and monitoring tasks, especially in psychiatric healthcare.
Hence, future research should integrate additional physiological signals (e.g., skin conductance, temperature, physical activity) and expand datasets to include diverse, representative populations and a broader spectrum of psychiatric disorders to explore HRV differences among psychiatric disorders and improve clinical insights. 
A multi-modal approach combining various data streams (e.g., eye-tracking, EEG) with XAI algorithms can offer a holistic understanding of physiological states, improving the accuracy of detecting psychiatric disorders. 
We aim to implement this approach using our high-performance computing infrastructure for large-scale, multi-dimensional data processing and analysis.

Beyond controlled studies, the ease of ambulatory HR measurement with wearables has greatly expanded interest in HRV for mental health, for example, stress and recovery can be tracked through smartphone apps paired with chest straps or smartwatches \cite{alugubelli2022wearable}. 
This continuous stream of data can complement traditional psychiatric assessments. For instance, a patient with bipolar disorder might wear an ECG sensor at home to continuously log HRV. If the data shows a sustained decline in vagal tone, which can be detected as a drop in nightly HRV or a rise in resting HRV measures, clinicians could be alerted that a depressive or manic shift is likely, prompting a proactive intervention. In schizophrenia, subtle signs of heightened autonomic arousal might indicate rising stress or early psychotic symptoms, and a wearable could flag these for further evaluation. Thus, wearable ECGs could extend the reach of psychiatric monitoring from the clinic to the patient's daily environment, capturing physiological changes associated with mental states that might otherwise go unnoticed.
We also underscore the importance of open-access and customizable HRV data collection in wearable devices. This highlights a potential area for development in consumer wearables that can balance the flexibility and security of HRV monitoring in both research and clinical applications.

\subsection{Toward Human-centered Contestable AI Systems in Healthcare: Limitations, Challenges and Future Works}
\subsubsection{Technical and LLMs' Medical Knowledge Limitations}
Integrating SAEs with contestable LLMs represents a significant step toward human-centered AI systems in healthcare. By creating a multi-layered verification framework, our approach addresses one of the fundamental challenges in clinical AI deployment (i.e., balancing the efficiency of automation with the necessity for appropriate human oversight). However, building truly effective contestable AI systems presents several challenges that must be addressed through ongoing research and development.

While our evaluation demonstrates a clear correlation between explanation discrepancies and prediction errors, determining optimal thresholds for flagging potentially unreliable predictions remains challenging. The threshold effect we observed (around 5-7 discrepancies) provides a useful starting point, but clinical deployment would require more robust calibration across diverse patient populations. Research in medical AI ethics reflects this tension, with some experts advocating context-dependent levels of explainability rather than one-size-fits-all standards \cite{freyer2024ethical}. Future work should explore adaptive thresholds that account for case-specific complexity.  

Despite impressive performance, current LLMs exhibit inconsistent medical knowledge. The variation in contestation effectiveness across different architectures highlights the limitations of their pre-trained knowledge about psychiatric disorders and HRV interpretation. The larger models generally perform better, but even $\texttt{gemma-3-instruct (27B)}$ fails to correct half of the incorrect predictions. For example, a study on psychiatric note analysis found that even when augmented with guidelines, LLMs \textit{lack a robust understanding of the meaning and nuances} of mental health codes \cite{boggavarapu2024evaluating}. This emphasizes the need for domain-specific optimization and more comprehensive incorporation of clinical expertise. For example, domain-tuned models like Google's Med-PaLM have shown significantly better alignment with medical consensus and fewer content omissions or biases than their general-purpose counterparts \cite{singhal2023large}.

\subsubsection{Challenges in Clinical Validation for Human-Centered Contestable AI System}
A challenge raised from our current study is how to conduct comprehensive clinical evaluations for a human-centered contestable AI system, which stems from the multidisciplinary expertise required for thorough assessment \cite{mansi2025legally}. 
Proper clinical validation necessitates collaboration with specialists who have dual expertise in both psychiatric disorders and cardiac signal interpretation, which is a relatively rare combination in clinical practice. 
Furthermore, meaningful evaluation must assess multiple aspects, including clinical utility, workflow integration, and long-term impact on treatment outcomes. Our preliminary technical validation with the HRV-ACC dataset provides encouraging results, but we recognize that full clinical implementation requires more extensive real-world testing across diverse patient populations, varying clinical settings, and different types of healthcare providers. This comprehensive evaluation framework aligns with emerging best practices in medical AI assessment, which emphasize contextual performance rather than technical metrics. 

We are currently developing partnerships with clinical institutions to facilitate this next phase of evaluation, which will incorporate both quantitative performance measures and qualitative feedback from healthcare providers and patients.
Our future work will also detail the comparative predictive performance and extend the applicability of the contestable LLMs system to different use cases (i.e., detection of atrial fibrillation and other conditions).

\subsubsection{Visions for Future Contestable AI Systems}
The vision of truly contestable AI systems extends beyond technical improvements to encompass broader considerations of agency, ethics, and human-AI interaction. 
Our work shows the potential of the contestable LLMs systems in providing an initial rationale for a diagnosis, and if the clinician is unconvinced or presents counter-evidence, the AI can delve deeper to justify or revise its conclusion. This aligns with the vision that machines should be able to assess the grounds for contestation given by humans and modify their decision-making if a valid issue is raised \cite{leofante2024contestable}.
A promising direction is adaptive explainability, where explanation style, complexity, and detail automatically adjust to match the specific context and user needs, moving beyond heuristic prompt engineering approaches. 
This could mean simplifying explanations for a junior clinician or providing more technical, detailed justifications for an expert specialist. It could also mean highlighting different content, for example, emphasizing pathophysiology for a physician versus explaining in lay terms for patient-facing contexts. Future research may explore methods for AI to gauge the clinician’s needs, which could be via user profiles or real-time interactions, and then tailor its explanations. Moreover, learning to explain frameworks might be developed, where models are trained not just to maximize accuracy but also to optimize explanation human-centered utility metrics, such as usefulness, plausibility, faithfulness, and fairness \cite{kong2024toward,nguyen2023towards}. By incorporating feedback from clinicians on which explanations were helpful or not, LLMs could refine their explanation and contestability strategies over time, which could improve efficiency (providing just enough detail) and effectiveness (ensuring the clinician understands the AI's reasoning) in everyday use. 

Future contestable AI should also employ multi-model agreement, where cross-checking answers with a second opinion from another model (or a simpler algorithmic rule) to see if there is consensus. In the context of patient vital signs or sensor data (e.g., continuous ECG monitors), real-time verification might involve signal processing algorithms validating an LLM's interpretation of the data. Research is needed on how to seamlessly weave these verification steps into the clinical workflow without causing delays. The end goal is an early warning system for AI errors: \textit{a contestable AI that not only explains itself, but also proactively points out why it might be wrong and shows external evidence for clinicians to verify.}

We also expect new governance mechanisms, such as hospital AI oversight committees or continuous monitoring of AI models in deployment to detect drifts or biases that could affect contestability (e.g., AI whose explanations degrade over time due to data shifts). 
Feedback loops in governance are equally important, where clinicians and patients should have avenues to report AI errors or problematic decisions, prompting model improvements or regulatory action if needed \cite{mansi2025legally}. 
Future contestable AI systems will need to operationalize guidelines. For instance, by logging all AI recommendations and clinician overrides to allow audit trails (supporting accountability), or by implementing safety constraints that require human confirmation for certain high-stakes AI actions.

To summarize, a multidisciplinary effort is required, involving human AI interaction researchers, clinicians, ethicists, and regulators, to ensure that the next generation of AI systems in healthcare is not only technically robust but also aligned with ethical norms, legal requirements, and the day-to-day realities of clinical care. Human-centered contestable AI will prioritize transparency, enable active clinician engagement, and maintain robust safeguards so that these tools truly enhance healthcare outcomes without eroding human agency.

%% file: sec/8_cls.tex
\section{Conclusion}\label{sec:conc}
Our research demonstrates that combining wearable ECG monitoring with human-centered contestable AI creates a viable pathway toward more objective and reliable psychiatric diagnosis. Heart2Mind successfully addresses several key challenges in current psychiatric assessment: the subjectivity of traditional diagnostic methods, the need for continuous monitoring capabilities, and the critical requirement for human oversight in AI-assisted healthcare decisions.
The MSTFT model's superior performance (91.7\% accuracy) validates the effectiveness of integrating multi-scale temporal and frequency features for analyzing cardiac biomarkers. More significantly, our self-adversarial explanations successfully identify potential model errors, with discrepancy counts serving as reliable indicators of prediction uncertainty. The contestable LLMs component ensures that clinical expertise remains central to the diagnostic process, achieving perfect accuracy in validating correct predictions while successfully challenging 50\% of erroneous ones.
This work has important implications for the future of psychiatric care. By enabling continuous, objective monitoring through consumer-grade wearables while maintaining clinical oversight through contestable AI, Heart2Mind offers a practical approach to early detection and monitoring of psychiatric disorders. The system's ability to capture subtle autonomic dysfunction patterns could lead to earlier interventions and more personalized treatment strategies.
Future work should address the integration of multiple physiological signals, expand datasets to include more diverse populations and psychiatric conditions, and explore domain-specific optimization of LLMs for enhanced clinical accuracy. As regulatory frameworks increasingly emphasize the need for explainable and contestable AI in healthcare, systems like Heart2Mind provide a foundation for developing trustworthy AI tools that augment rather than replace clinical expertise, ultimately improving patient outcomes in psychiatric care.

%% file: sec/9_appx.tex
\section*{Acknowledgment}
This study is supported by the UNB-FCS Startup Fund (22–23 START UP/H CAO). It is also funded in part by the National Research Council of Canada’s Aging in Place Challenge program.